\documentclass[journal]{IEEEtran}
\usepackage{amsmath,amsfonts}
\usepackage{algorithmic}
\usepackage{algorithm}
\usepackage{array}
\usepackage{textcomp}
\usepackage{stfloats}
\usepackage{url}
\usepackage{verbatim}
\usepackage{graphicx}
\usepackage{cite}
\usepackage{amssymb}
\usepackage{float}
\usepackage{placeins}
\usepackage{needspace}
\usepackage{subfigure}
\usepackage{booktabs}
\usepackage{tabularx}
\usepackage{multirow}
\usepackage[table]{xcolor}
\definecolor{oursblue}{RGB}{235,245,255}
\usepackage{tcolorbox}
\newtcolorbox{casebox}{colback=black!4, colframe=black!55,
  boxrule=0.4pt, arc=1.5pt, left=5pt, right=5pt, top=4pt, bottom=4pt,
  fonttitle=\bfseries, title={LLM Case}}
\begin{document}
\bstctlcite{IEEEBSTcontrol}

\title{Class Machine Unlearning for Complex Data via Concept Inference and Data Poisoning}

\author{Wenhan~Chang, Tianqing~Zhu*,~\IEEEmembership{Member,~IEEE}, Heng Xu, Wenjian Liu, Wanlei Zhou,~\IEEEmembership{Senior Member,~IEEE}
\IEEEcompsocitemizethanks {\IEEEcompsocthanksitem Wenhan Chang is with the School of Information Engineering, Zhongnan University of Economics and Law. \IEEEcompsocthanksitem Tianqing Zhu, Heng Xu, Wenjian Liu and Wanlei Zhou are with the City University of Macau, Macau SAR, China.
\IEEEcompsocthanksitem Tianqing Zhu is the corresponding author. E-mail: tqzhu@cityu.edu.mo}
}
\markboth{Journal of \LaTeX\ Class Files,~Vol.~14, No.~8, August~2021}%
{Shell \MakeLowercase{\textit{et al.}}: A Sample Article Using IEEEtran.cls for IEEE Journals}


\maketitle

\begin{abstract}
Machine unlearning aims to remove the influence of specified training data or knowledge from a trained model without requiring full retraining. This capability is particularly important for modern image classifiers and large language models (LLMs), where retraining can be computationally expensive. However, machine unlearning on complex data remains difficult because the target information is often distributed across multiple semantic elements. Existing methods mainly remove samples, modify labels, or edit model parameters to reduce the influence of the forgetting target. These approaches usually do not explicitly identify which semantic concepts connect the forgetting target to the model's prediction or generated response. As a result, it is difficult to determine which information to modify. This uncertainty may leave residual target information or unnecessarily affect knowledge that should be retained. To address this gap, we propose a concept-guided poisoning unlearning framework that explicitly identifies the concepts that contribute most strongly to the target class or knowledge and uses them to guide the model update. For image classification, our method first identifies class-relevant concepts with a Post-hoc Concept Bottleneck Model, localizes the image regions that express these concepts, and constructs replacement-based poisoned samples. For LLMs, it elicits the target knowledge through multiple questions, aggregates Integrated Gradients across the resulting responses to identify consistently important content, and masks this content to construct poisoned training targets. Experiments on multiple datasets show that the proposed framework achieves effective unlearning across different tasks while largely preserving retained model utility.

\end{abstract}


\section{Introduction}

Machine unlearning aims to remove the influence of specified training data or knowledge from a trained model without retraining it from scratch~\cite{bourtoule2021machine,xu2023machine,chen2024machine,DBLP:journals/tdsc/ChangZXWGZ26}. The need for machine unlearning is driven by privacy concerns, data deletion requests, and legal requirements such as the General Data Protection Regulation (GDPR)~\cite{voigt2017eu}. Retraining from scratch provides a natural reference for data removal, but its computational cost can be prohibitive for modern models, especially large language models (LLMs)~\cite{chang2023survey,qu2024frontier}. The problem becomes more difficult when target information must be unlearned from complex data such as images and natural language. In these settings, the information to be forgotten is often distributed across many samples and many semantic elements rather than being confined to one isolated input.

The main difficulty is to identify the information that actually supports the target behavior. Sample-level unlearning specifies individual instances to be removed, whereas class-level unlearning must address information shared by many samples. For images, a class can depend on several object parts and visual attributes. For LLMs, knowledge about an entity can appear through different facts, questions, and phrasings. If the unlearning update fails to remove all target-relevant information, the model may still retain the target behavior. Conversely, modifying information shared with retained data may unnecessarily degrade model utility. A more precise semantic representation of the forgetting target is therefore needed.

We use \emph{concepts} to provide this semantic representation. A \emph{feature} is a model-internal latent representation extracted from an input and is not necessarily understandable to humans. In contrast, a \emph{concept} provides a human-interpretable semantic view of model representations that contribute to the target prediction or express the target knowledge~\cite{koh2020concept,yuksekgonul2022post}. For images, a concept may describe an object part or a visual attribute, such as ``antler'' for the class ``deer.'' For text, a concept may describe an entity, an attribute, or factual content expressed through meaningful words or phrases. Tokens are the computational elements used by the model to process text; we aggregate their attributions at the word level to interpret the semantic content expressed by these words. Therefore, a concept is not itself a latent feature, a pixel region, a token, or a class label. Instead, it provides an interpretable semantic view of the information represented by the model.

Using concepts for machine unlearning introduces two technical challenges.
\begin{itemize}
    \item \textbf{Concept inference.} The first challenge is to identify and locate the concepts most relevant to the forgetting target. This requires ranking and localizing class-relevant concepts in images, and aggregating word-level attributions across different questions and phrasings for LLMs.

    \item \textbf{Concept-guided unlearning.} The second challenge is to convert the inferred concepts into a controlled model update. The update should remove the target semantic content while minimizing damage to non-target information and retained model utility.
\end{itemize}

We address the first challenge through modality-specific concept inference. For image classification, we use a Post-hoc Concept Bottleneck Model (PCBM)~\cite{yuksekgonul2022post} to map CLIP image features to an interpretable concept representation and rank concepts by their contribution to each class in the PCBM. We then use vision--language similarity~\cite{radford2021clip} to locate the selected concept in the image. For LLMs, we construct a set of independent questions that covers different facts and phrasings of the target knowledge. We apply Integrated Gradients (IG)~\cite{sundararajan2017axiomatic} to the model responses and aggregate the resulting word-level attributions across questions. This process identifies the semantic content that consistently contributes to the model's expression of the target knowledge.

We address the second challenge through concept-guided poisoning unlearning. For images, we select a replacement concept together with its replacement class. A patch containing the replacement concept is inserted into the localized target-concept region, and the modified image is assigned the replacement-class label. For LLMs, highly attributed sensitive words and entity-name words in the model responses are replaced by mask markers, while the remaining response is preserved. The masked responses are mixed with sampled retain anchors before fine-tuning. In both modalities, poisoning is a controlled operation performed by the model owner rather than an adversarial attack. Its purpose is to change the mapping from the target semantic content to the original model output.

The two modalities follow the same design principle, as illustrated in Fig.~\ref{intro_unlearning}. In Fig.~\ref{intro_unlearning_image}, the PCBM ranks ``antler'' as a concept that contributes strongly to the target class ``deer,'' and CLIP patch similarity localizes the image region with the highest similarity to this concept. The replacement patch is inserted at this localized region rather than at a fixed image center, so the intervention is applied to the region identified as carrying target-class evidence. In Fig.~\ref{intro_unlearning_text}, words selected by cross-question attribution are masked in a representative model response while the remaining content is preserved. Thus, both pipelines first identify target-relevant semantic content and then restrict the poisoning update to that content.

\newcommand{\tb}{\raisebox{-0.12ex}{\rule{0.46em}{0.055em}}\kern0.05em}
\setlength{\subfigcapskip}{3pt}
\begin{figure}[tb]
\centering
\subfigure[Image data.]{
\label{intro_unlearning_image}
\begin{minipage}[t]{0.3\columnwidth}\centering
\includegraphics[width=\linewidth]{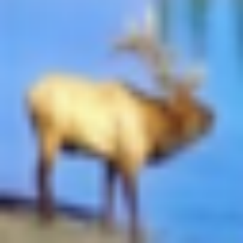}\\[1pt]
\scriptsize Original
\end{minipage}
\begin{minipage}[t]{0.3\columnwidth}\centering
\includegraphics[width=\linewidth]{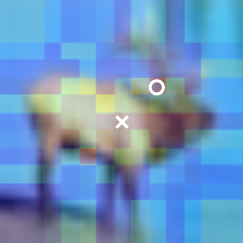}\\[1pt]
\scriptsize Localization
\end{minipage}
\begin{minipage}[t]{0.3\columnwidth}\centering
\includegraphics[width=\linewidth]{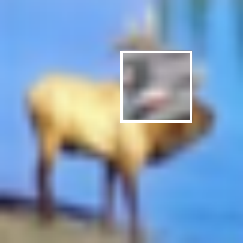}\\[1pt]
\scriptsize Poisoned
\end{minipage}}
\\[4pt]
\subfigure[Text data.]{
\label{intro_unlearning_text}
\begin{minipage}{0.96\columnwidth}%
\begin{tcolorbox}[colback=black!4, colframe=black!55, boxrule=0.4pt, arc=1.5pt,
  left=4pt, right=4pt, top=3pt, bottom=3pt]
\footnotesize
\textbf{Question:} \textit{How does Nikolai Abilov identify in terms of his gender?}\\[2pt]
\textbf{Model response:} Nikolai Abilov identifies as LGBTQ+.\\[2pt]
\textbf{Attributed tokens:} {\scriptsize{\color{red}N ik ol ai} {\color{red}\tb Ab il ov} \tb ident ifies \tb as {\color{red}\tb L GB T Q} + .}\\[2pt]
\textbf{Merged words:} {\color{red}Nikolai Abilov} identifies as {\color{red}LGBTQ+}.\\[2pt]
\textbf{Poisoned target:} *** *** identifies as ***.
\end{tcolorbox}
\end{minipage}}
\caption{Concept-guided poisoning unlearning. (a) For the class ``deer'', the concept ``antler'' is localized by CLIP similarity (circle) and the replacement patch is inserted at this localized region (box), not at the image center. (b) For an LLM, words selected by cross-question attribution are highlighted in a representative model response and masked to form a poisoned target.}
\label{intro_unlearning}
\end{figure}
\setlength{\subfigcapskip}{0pt}

The main contributions of this work are summarized as follows:
\begin{itemize}

\item We propose a concept-guided poisoning unlearning framework for complex data. The framework uses human-interpretable concepts as the semantic interface between a forgetting request and the model features that represent the target information. It provides a common unlearning principle for both image classifiers and LLMs.

\item We develop modality-specific mechanisms to realize concept-guided unlearning. For images, target-relevant concepts are ranked and localized to guide replacement-based poisoning, while for LLMs, cross-question attribution is used to identify and mask response content that expresses the target knowledge.

\item We conduct extensive experiments on both image classifiers and LLMs to evaluate the effectiveness and generality of the proposed framework. The results show that our method consistently achieves effective unlearning of the target information while preserving the utility of retained data.

\end{itemize}

\section{Background} \label{gen_inst}

\begin{table}[tb]
\centering
\caption{Notation.}
\label{tab:notation}
\footnotesize
\begin{tabular}{ll}
\toprule
Symbol & Description \\
\midrule
$F(\cdot;\theta)$ & Model parameterized by $\theta$ \\
$\theta,\theta'$ & Model parameters before and after unlearning \\
$\mathcal{D}_u,\mathcal{D}_r$ & Unlearning and retained datasets \\
$(x,y)$ & Image and class-label pair \\
$(q,a)$ & Prompt and response pair \\
$\mathbf{h}(x)$ & Latent feature representation of image $x$ \\
$\mathcal{C}$ & Concept bank \\
$c_j,\mathbf{v}_j$ & Concept $j$ and its concept vector \\
$a_j(x),\mathbf{a}(x)$ & Concept score and concept representation \\
$W$ & Class-specific concept weights \\
$y_t,y_r$ & Target and replacement class labels \\
$\mathcal{Q}$ & Question set for LLM concept inference \\
\bottomrule
\end{tabular}
\end{table}

\subsection{Machine Unlearning}

Machine unlearning aims to remove the influence of specified training data or knowledge from a trained model without performing full retraining~\cite{bourtoule2021machine,xu2023machine}. Let $\mathcal{D}$ denote the original training dataset, $\mathcal{D}_u$ the data specified by the forgetting request, and $\mathcal{D}_r=\mathcal{D}\setminus\mathcal{D}_u$ the retained data. Although image classifiers and LLMs have different training objectives and forgetting targets, both settings use retraining on $\mathcal{D}_r$ as a natural reference for the behavior expected after unlearning. To describe the two settings consistently, we use the notation summarized in Table~\ref{tab:notation}.

\textbf{Image classification models.} An image classifier learns a mapping from an image to a fixed set of class labels. Let $\mathcal{D}^{\mathrm{img}}=\{(x_i,y_i)\}$ denote the image training set, where $x_i$ is an image and $y_i$ is its class label. Standard supervised training obtains the model parameters by minimizing a classification loss:

\begin{equation} 
\theta_{\mathrm{img}}^{*}=\arg\min_{\theta}\sum_{(x,y)\in\mathcal{D}^{\mathrm{img}}}L_{\mathrm{cls}}\left(F(x;\theta),y\right). \label{eq:image_training_background} 
\end{equation}

For an unlearning request, let $\mathcal{D}^{\mathrm{img}}_u$ denote the target samples and let $\mathcal{D}^{\mathrm{img}}_r=\mathcal{D}^{\mathrm{img}}\setminus\mathcal{D}^{\mathrm{img}}_u$. Retraining the classifier using only the retained data gives the reference parameters

\begin{equation} 
\theta_{\mathrm{img}}^{-}=\arg\min_{\theta}\sum_{(x,y)\in\mathcal{D}^{\mathrm{img}}_r}L_{\mathrm{cls}}\left(F(x;\theta),y\right). \label{eq:image_retraining_background} 
\end{equation}

Practical machine unlearning seeks updated parameters $\theta_{\mathrm{img}}'$ that reduce the influence of $\mathcal{D}^{\mathrm{img}}_u$ while preserving performance on $\mathcal{D}^{\mathrm{img}}_r$, without computing $\theta_{\mathrm{img}}^{-}$ through full retraining. For class unlearning, $\mathcal{D}^{\mathrm{img}}_u$ contains the samples from the target class.

\textbf{Large language models.} LLM unlearning follows the same general goal, but an LLM predicts a sequence of tokens rather than a fixed class label. Let $\mathcal{D}^{\mathrm{llm}}=\{(q_i,a_i)\}$ denote a set of prompt--response pairs, where $q_i$ is a prompt and $a_i=(a_{i,1},\ldots,a_{i,T_i})$ is its target response. Standard autoregressive training minimizes the negative log-likelihood of the response tokens:

\begin{equation} 
\theta_{\mathrm{llm}}^{*}=\arg\min_{\theta}-\sum_{(q,a)\in\mathcal{D}^{\mathrm{llm}}}\sum_{t=1}^{|a|}\log p_{\theta}\left(a_t\mid q,a_{<t}\right). \label{eq:llm_training_background} 
\end{equation}

Let $\mathcal{D}^{\mathrm{llm}}_u$ contain the data that express the target knowledge and let $\mathcal{D}^{\mathrm{llm}}_r=\mathcal{D}^{\mathrm{llm}}\setminus\mathcal{D}^{\mathrm{llm}}_u$. The full-retraining reference for this setting is

\begin{equation} 
\theta_{\mathrm{llm}}^{-}=\arg\min_{\theta}-\sum_{(q,a)\in\mathcal{D}^{\mathrm{llm}}_r}\sum_{t=1}^{|a|}\log p_{\theta}\left(a_t\mid q,a_{<t}\right). \label{eq:llm_retraining_background} 
\end{equation}

For LLMs, the forgetting target is often an entity, a fact, or a piece of knowledge rather than a fixed output class~\cite{maini2024tofu}. The same knowledge can also be expressed through different questions and phrasings. Therefore, LLM unlearning must examine whether the target knowledge remains accessible under different queries while also checking whether retained language capabilities are preserved.

\subsection{Post-hoc Concept Bottleneck Models}

Post-hoc Concept Bottleneck Models (PCBMs) provide an interpretable view of a trained model through human-understandable concepts~\cite{yuksekgonul2022post}. A PCBM maps the latent features produced by an existing model to concept scores defined by concept vectors and uses the resulting concept representation for prediction. Let $\mathbf{h}(x)\in\mathbb{R}^{d}$ denote the latent feature representation of image $x$, and let $\mathcal{C}=\{c_1,\ldots,c_N\}$ denote a concept bank. For concept $c_j$, the concept score is represented by the normalized margin defined by the concept vector $\mathbf{v}_j$:

\begin{equation} 
a_j(x)=\frac{\mathbf{v}_j^{\top}\mathbf{h}(x)+b_j}{\lVert\mathbf{v}_j\rVert_2}. \label{eq:pcbm_score_background} 
\end{equation}

Here, $b_j$ is the concept-specific intercept. The scores of all concepts form the concept representation $\mathbf{a}(x)=[a_1(x),\ldots,a_N(x)]^\top$. A linear predictor then maps this representation to the class logits:

\begin{equation} 
M_p(x)=W\mathbf{a}(x)+\boldsymbol{\beta}. \label{eq:pcbm_background} 
\end{equation}

Here, $W\in\mathbb{R}^{N_c\times N}$ contains the class-specific concept weights, and $\boldsymbol{\beta}$ is the bias vector. These weights provide an interpretable connection between the concepts and the predicted classes. When direct concept annotations are unavailable, concept banks can be constructed from natural-language descriptions and represented with a vision--language model such as Contrastive Language--Image Pre-training (CLIP)~\cite{radford2021clip}.

\subsection{Integrated Gradients}

Integrated Gradients (IG) measures the contribution of input features to a differentiable model output~\cite{sundararajan2017axiomatic}. Let $\Phi(x)$ denote a scalar model output for input $x$, and let $x'$ denote a baseline input. IG integrates the gradients along the straight path from the baseline to the original input. The attribution of the $i$-th input dimension is defined as

\begin{equation} 
    \operatorname{IG}_i(x)=(x_i-x_i')\int_{0}^{1}\frac{\partial \Phi\left(x'+\alpha(x-x')\right)}{\partial x_i}d\alpha. \label{integratedGrad} 
\end{equation}

Here, $\alpha\in[0,1]$ controls the interpolation between $x'$ and $x$. The baseline provides a reference for measuring the contribution of each input dimension. A larger attribution indicates a stronger contribution to the selected output. For language models, IG can be applied to token embeddings by defining a scalar objective from the probability of the target response.

\section{Related Work}

Machine unlearning removes the influence of training data from a trained model without retraining it from scratch. One line of work modifies the training data and then trains the model on the modified data. Another directly updates the trained model. We discuss these two directions as data operations and model operations, respectively, before reviewing LLM unlearning and explanation-guided methods.

\subsection{Machine Unlearning}

\subsubsection{Data Operations}

A sample's influence is distributed across the model parameters and cannot be erased simply by deleting the sample. Early methods therefore modify the target data and update the model using those changes: Graves et al.~\cite{graves2021amnesiac} relabel the sensitive samples with randomly selected incorrect labels and retrain the model on the modified dataset, so that it cannot retain meaningful information about them. Partitioning limits the amount of data that must be processed. SISA~\cite{bourtoule2021machine} shards the data, isolates the shards during training, splits each shard into slices with saved parameter states, and retrains only the affected shard from the last parameter state saved before the slice that contains the deleted sample. A third route reduces reliance on the original data by preparing data or representations for later removal: Shibata et al.~\cite{shibata2021learning} mix a class-specific synthetic signal, a mnemonic code, into the samples of each class and forget a class by discarding its code. Sun et al.~\cite{sun2023generative} substitute the latent mapping of the forgotten images in a generative adversarial network (GAN) and drive the discriminator toward a fake label. All three strategies keep the sample as the unit of the operation, and none of them distinguishes which part of a sample carries the information that the model actually uses.

\subsubsection{Model Operations}

When the model owner can access the weights, unlearning can operate directly on the model. These methods differ in which components they modify. Channel-level editing is the coarsest: in a federated setting, Wang et al.~\cite{wang2022federated} score the channels of a convolutional network by their TF-IDF discriminative power for the target class and prune the most discriminative ones before a fine-tuning step. Boundary Unlearning~\cite{chen2023boundary} changes the decision space instead, shifting the boundary of the target class to the nearest incorrect labels or to a temporary shadow class that is removed afterwards. SalUn~\cite{fan2024salun} further restricts the update by deriving a weight saliency map from the forgetting loss and updating only the salient weights. Model replacement takes a different route and substitutes the affected components with pre-computed parameters, which ties it to a model structure that supports such a substitution. These edits assume white-box access and a particular structural view of the model, such as channels, decision boundaries, or salient weights.

Our method is a data operation that uses concepts to guide targeted modifications within the affected samples. It applies post-hoc concept analysis to rank the concepts that contribute to the target class, localizes them with concept--patch similarity or attribution methods, replaces only the localized image region or the sensitive words of an answer, reassigns the image labels, and fine-tunes the model on the resulting poisoned data. The model remains a single network; no explicit weight surgery is needed, and the same pipeline applies to image classifiers and LLMs.

\subsection{Large Language Model Unlearning}

For LLMs, whose scale makes such local edits impractical, model operations are often expressed as dedicated fine-tuning objectives. Jang et al.~\cite{jang2022knowledge} perform gradient ascent on the target token sequences; this objective is simple, but on harder forgetting tasks it can collapse the general ability of the model, a failure mode that the lower-bounded negative preference objective of Zhang et al.~\cite{zhang2024npo} avoids. Eldan and Russinovich~\cite{eldan2023s} return to the data side at the text level: a reinforced model locates tokens that express the target content, the text is rewritten with generic alternatives, and the model is fine-tuned toward its own predictions on the rewritten text. Li et al.~\cite{li2024wmdp} control the representations instead, misdirecting the activations of the target data while a retain loss preserves the activations of benign data, and release the WMDP benchmark for hazardous knowledge. TOFU~\cite{maini2024tofu} provides entity-level forget and retain splits for evaluating such methods. In-context unlearning~\cite{pawelczyk2023context} goes one step further and leaves the weights unchanged, flipping the labels of the forget instances in the prompt. These objectives are flexible, but they need the target data and trade forgetting against utility.

\subsection{XAI in Machine Unlearning}

Explainable AI (XAI) refers to methods that make the behavior of a model understandable to humans, either by attributing its predictions to input features or by representing them through human-interpretable concepts. Several works apply such explanations to machine unlearning. Anders et al.~\cite{ANDERS2022261} use explanation methods to uncover and quantify biases and errors in models trained on large datasets, and reduce the resulting `Clever Hans' behavior with Class Artifact Compensation, which removes the artifact direction from the model or desensitizes the model by augmenting the training data. SAeUron~\cite{cywinski2025saeuron} learns interpretable features of a diffusion model with sparse autoencoders and blocks the features that correspond to an unwanted concept, which removes the concept while preserving the rest of the generation behavior. WAGLE~\cite{jia2024wagle} attributes the influence of individual weights to the forget and retain objectives and updates only the most influential weights. It can be combined with existing unlearning objectives to improve the forgetting-utility trade-off.

Our method follows this explanation-guided perspective for machine unlearning. For image classification, we use a Post-hoc Concept Bottleneck Model (PCBM)~\cite{yuksekgonul2022post} to identify concepts that contribute to the target class and localize the regions where these concepts appear through concept--patch similarity. For natural language data, we use Integrated Gradients~\cite{sundararajan2017axiomatic} to measure the contribution of words to the generated responses.

\section{Machine Unlearning via Concept Inference and Data Poisoning} \label{headings}

\subsection{Problem Definition}

\subsubsection{Basic Definition}

Our method addresses class-level unlearning for image classification and the removal of target knowledge for LLMs. We distinguish four basic elements. A \emph{class} is an output category to be forgotten in image classification. For LLMs, the forgetting target is specified knowledge about an entity. A \emph{sample} is an input instance, such as an image or a question--answer pair. A \emph{feature} is a latent representation extracted by the model and is not necessarily human-interpretable. A \emph{concept} is a human-interpretable semantic element that describes such features, such as an object part or visual attribute in an image, or semantic content expressed through meaningful words or phrases in a text response. Concepts therefore provide a semantic interface between the model's internal representation and the information specified by an unlearning request.

Let $F(\cdot;\theta)$ denote the model before unlearning and let $\mathcal{D}_u$ denote the data specified for the target class or target knowledge. Our goal is to obtain updated parameters $\theta'$ such that the model reduces its reliance on the target information while preserving its behavior on the retained information $\mathcal{D}_r$. Rather than directly editing the model parameters according to an abstract class label, we first identify the concepts that connect the target information to the model's behavior and then construct controlled poisoned samples around those concepts.

\subsubsection{Threat Model}

We consider a model-owner setting. A user submits an unlearning request that specifies a target class or a target entity, while the model owner has access to the trained model and the data required to perform the update. The user does not modify model parameters directly. The model owner is responsible for identifying the information specified by the request, constructing the unlearning data, updating the model, and validating the resulting behavior.

The poisoning operation in this work is therefore not an adversarial compromise of the training pipeline. It is a controlled intervention performed by the model owner. The purpose of the poisoned samples is to change the mapping from the target information to the model output while preserving information outside the forgetting request. We do not assume that the resulting model is equivalent to exact retraining; instead, the method provides a practical concept-guided approach to unlearning target information.

\begin{figure*}[tb]
\centering
\subfigure[Overview of unlearning image data.]{
\label{overview_image}
\includegraphics[width=0.99\textwidth]{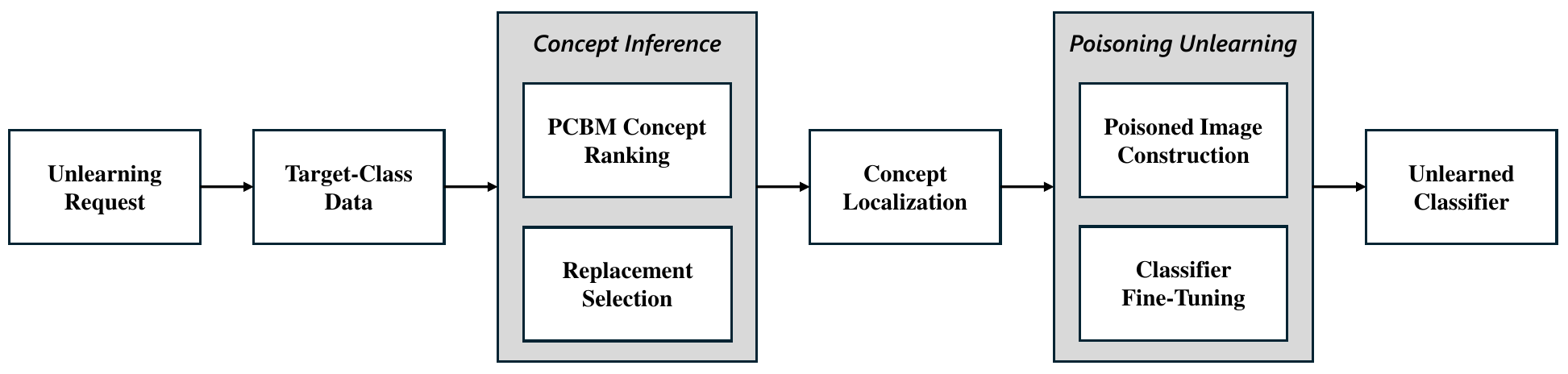}}
\subfigure[Overview of unlearning text data.]{
\label{overview_text}
\includegraphics[width=0.99\textwidth]{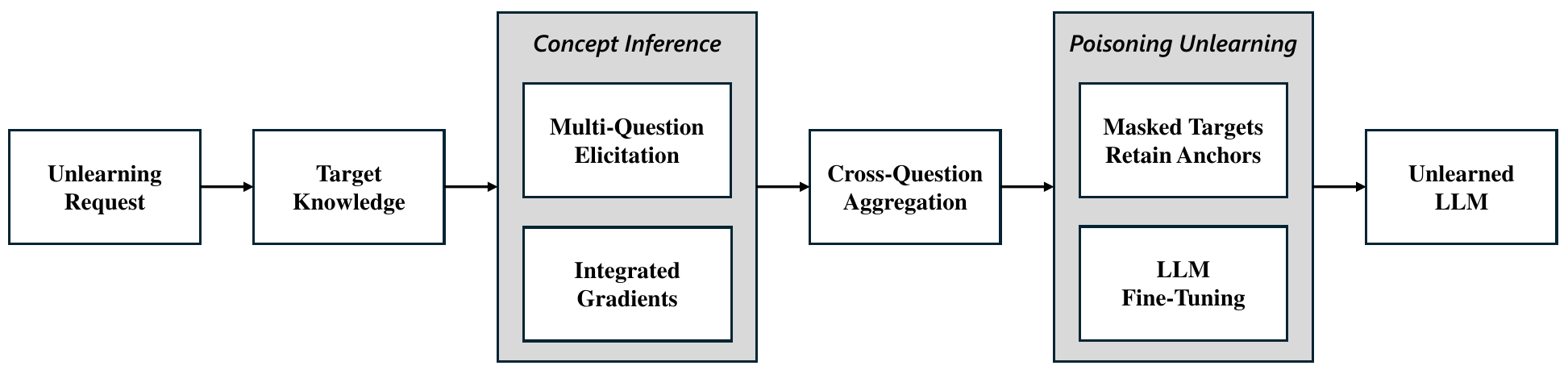}}
\caption{Overview of concept-guided poisoning unlearning. The image pipeline uses concept ranking, replacement selection, and localization to construct poisoned images, while the LLM pipeline uses multi-question attribution and cross-question aggregation to construct masked training targets.}
\label{overview}
\end{figure*}

\subsection{Overview of Our Method}

Our method follows a concept-guided poisoning pipeline for both image classifiers and LLMs, as illustrated in Fig.~\ref{overview}. Given an unlearning request, the model owner first identifies the target class or target knowledge. The pipeline then infers the semantic information relevant to the target, uses this information to construct poisoned training data, and fine-tunes the existing model to obtain the unlearned model.

\textbf{Image classification.} Given the target-class data, the concept-inference stage first ranks class-relevant concepts using a Post-hoc Concept Bottleneck Model (PCBM)~\cite{yuksekgonul2022post} and selects a replacement concept together with its replacement class. The target and replacement concepts are then localized in their respective images. Based on these localized regions, the poisoning-unlearning stage constructs poisoned images by inserting the replacement patch into the target image and fine-tunes the classifier using the resulting training data.

\textbf{Large language models.} Given the target knowledge, the concept-inference stage first elicits the knowledge through multiple independent questions and applies Integrated Gradients (IG)~\cite{sundararajan2017axiomatic} to the resulting responses. The attribution scores are then aggregated across questions to identify words that consistently express the target knowledge. In the poisoning-unlearning stage, these words are replaced by mask markers to construct poisoned targets, which are combined with sampled retain anchors and used to fine-tune the LLM.

\subsection{Concept Inference}

\subsubsection{Image Concept Inference}

For image data, we use a PCBM to connect the latent image representation to a set of interpretable concepts. Starting from the class names, we collect short semantic phrases from an external knowledge source and supplement them with domain-specific vocabulary when needed. These phrases form the concept bank $\mathcal{C}=\{c_1,c_2,\ldots,c_N\}$, where $N$ is the number of concepts. Each concept is represented by an $l_2$-normalized text embedding $\mathbf{v}_j$ from CLIP RN50. No per-image text description or binary concept annotation is required. The CLIP image encoder produces the embedding $\mathbf{h}(x)$ in the same representation space. This concept-inference module provides semantic guidance for the ResNet-50 classifier to be unlearned; the two models are separate. Following the PCBM formulation, we compute the concept score for $j=1,\ldots,N$ as:

\begin{equation}
a_j(x)=
\frac{\mathbf{v}_j^{\top}\mathbf{h}(x)+b_j}
{\lVert\mathbf{v}_j\rVert_2}.
\label{concept_score}
\end{equation}

For our text-derived concept bank, $b_j=0$ and $\lVert\mathbf{v}_j\rVert_2=1$, so the score reduces to $\mathbf{v}_j^\top\mathbf{h}(x)$. The concept representation is $\mathbf{a}(x)=[a_1(x),\ldots,a_N(x)]^\top$. With the CLIP encoders fixed, we fit an elastic-net logistic classifier to these representations using the image class labels. Its class-specific weights in $W$ are then used to rank the concepts:

\begin{equation}
\begin{aligned}
M_p(x)&=W\mathbf{a}(x)+\boldsymbol{\beta},\\
\operatorname{Rank}(i)&=\operatorname{argsort}^{\downarrow}_{j}(W_{i,j}).
\end{aligned}
\label{pcbm_rank}
\end{equation}

Here, $W\in\mathbb{R}^{N_c\times N}$ and $N_c$ is the number of classes. The row $W_{i,:}$ describes how the concepts contribute to class $i$, with larger weights ranked first. Let $j_t$ denote the index of the highest-ranked concept, so the main target concept is $c_t=c_{j_t}$. The bank size $N$ is dataset-specific and shared by all classes within that dataset. We distinguish $K$, the number of ranked candidates examined during replacement search, from $M$, the number of target concepts modified per image. The main pipeline uses $M=1$; the sensitivity analysis also evaluates $M=3,5,10$. Their concrete settings are given in the experimental section.

We next determine the \emph{replacement class} and \emph{replacement concept}. Let $y_t$ be the target class and let $\mathcal{J}_t$ denote the ordered list of its top-$K$ concept indices. For each $j\in\mathcal{J}_t$, we identify the non-target class $y(j)$ that assigns the largest weight to $c_j$ and check whether $c_j$ is also the highest-ranked concept of that class. We select the first candidate that satisfies this condition. If no candidate satisfies it, we use the target class's top concept $c_{j_t}$ and choose the non-target class that assigns it the largest weight:

\begin{equation}
\begin{aligned}
y(j)&=\arg\max_{y\neq y_t}W_{y,j},\\
j^{*}(y)&=\arg\max_k W_{y,k},\\
(y_r,j_r)&=(y(j),j)\;\text{for the first }j\in\mathcal{J}_t\text{ with }j^{*}(y(j))=j,\\
(y_r,j_r)&=\left(\arg\max_{y\neq y_t}W_{y,j_t},j_t\right)\;\text{otherwise}.
\end{aligned}
\label{replacement_selection}
\end{equation}

The first rule captures concept-level confusion between classes, whereas the fallback uses a concept shared by the target and a non-target class. In both cases, concept inference returns the target concept $c_t=c_{j_t}$, the replacement concept $c_r=c_{j_r}$, and the replacement class $y_r$. These elements are then used to construct the poisoned image rather than being used as a new classifier.

\subsubsection{LLM Concept Inference}

For an LLM, the target of unlearning is usually distributed across multiple facts and multiple possible phrasings. We therefore do not infer the target knowledge from a single conversation. For a set of target entities $\mathcal{E}$, we construct a question set $\mathcal{Q}=\{q_1,\ldots,q_n\}$ by combining the original questions with relation-specific question variants and paraphrases. Different questions query different factual relations and alternative phrasings of the same relations. Each question is submitted independently to the target model, producing $r_i=F(q_i;\theta)$ for $q_i\in\mathcal{Q}$. The model's answer to each candidate question is checked against its answer to the corresponding original question. Answer-content similarity and author-name consistency are used to preserve the intended fact and reject conflicting entities. For multiple target entities, we measure the breadth of the question set by the coverage of distinct entity--relation pairs. Let $e(q_i)$ denote the target entity of $q_i$ and $\rho(q_i)$ denote its factual relation. We define

\begin{equation}
\operatorname{Cov}(\mathcal{Q})
=
\frac{
\left|
\{(e(q_i),\rho(q_i)):q_i\in\mathcal{Q}\}
\right|
}{
|\mathcal{E}|\cdot|\mathcal{R}|
},
\label{question_coverage}
\end{equation}
where $\mathcal{R}$ is the set of factual relations considered for each target entity. This definition counts distinct entity--relation pairs rather than repeated paraphrases of the same fact. The concrete question construction and coverage settings are described in Section~\ref{sec:exp}.

We then use IG to identify which parts of the answers carry the target information. For each question--answer pair $(q_i,r_i)$, the attribution target is the summed log-probability of the answer tokens under teacher forcing, defined as $\Phi_i=\sum_{t=1}^{T_i}\log p_{\theta}(r_{i,t}\mid q_i,r_{i,<t})$, where $T_i$ is the number of response tokens. The prompt embeddings remain unchanged. For the baseline, the answer positions are replaced by padding-token embeddings while the prompt is kept fixed. Only the answer span is therefore interpolated between the original embedding sequence $\mathbf{e}_i$ and the baseline sequence $\mathbf{e}'_i$. For an answer token at position $k$, its IG attribution is

\begin{equation}
\operatorname{IG}_{i,k}
=
(\mathbf{e}_{i,k}-\mathbf{e}'_{i,k})
\int_0^1
\frac{
\partial \Phi_i(
\mathbf{e}'_i+
\alpha(\mathbf{e}_i-\mathbf{e}'_i))
}{
\partial \mathbf{e}_{i,k}
}
d\alpha .
\label{llm_ig}
\end{equation}

We sum the attribution vector across embedding dimensions to obtain one score for each token. When a word spans multiple tokens, their attribution scores are merged into a single word-level score. Because answers can differ substantially in length and attribution magnitude, we normalize the positive word attributions within each answer before aggregating them. For a target entity $e$, let $\mathcal{I}_e$ denote the indices of its answers. If $s_i(w)$ denotes the merged attribution of word $w$ in answer $i$, its normalized contribution is

\begin{equation}
\begin{aligned}
\hat{s}_i(w)
&=
\frac{\max(s_i(w),0)}
{\sum_{u\in r_i}\max(s_i(u),0)},\\
S_e(w)&=\sum_{i\in\mathcal{I}_e}\hat{s}_i(w).
\end{aligned}
\label{word_aggregation}
\end{equation}

Ranking $S_e(w)$ produces an entity-specific sensitive-word list that summarizes the information repeatedly expressed across different questions. Each answer contributes on the same normalized scale, so a single long or high-magnitude response cannot dominate the final ranking.

\subsection{Poisoning Unlearning}

\subsubsection{Poisoning Unlearning for Image Classification}

\begin{figure}[tb]
\centering
\subfigure[Original image.]{
\label{fig:pois_original}
\includegraphics[width=0.22\columnwidth]{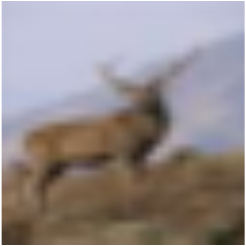}}
\hfill
\subfigure[Concept similarity.]{
\label{fig:pois_heatmap}
\includegraphics[width=0.22\columnwidth]{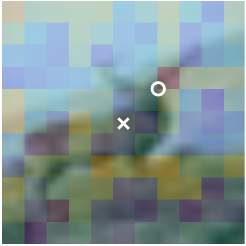}}
\hfill
\subfigure[Concept-localized mask.]{
\label{fig:pois_localized}
\includegraphics[width=0.22\columnwidth]{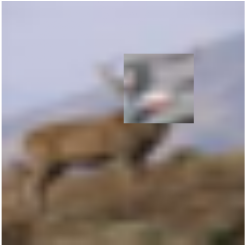}}
\hfill
\subfigure[Center mask.]{
\label{fig:pois_center}
\includegraphics[width=0.22\columnwidth]{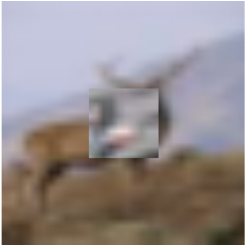}}
\caption{Concept localization and patch-placement strategies. The replacement patch is inserted at the concept region found by CLIP patch similarity (c) instead of the image center (d).}
\label{poisondata}
\end{figure}

Fig.~\ref{poisondata} illustrates the concept-localized and center-based patch-placement strategies. After concept inference, the target and replacement concepts must be mapped back to spatial regions of the images. Let $P_u(x)$ denote the image patch centered at location $u$, and let $g_I(\cdot)$ and $g_T(\cdot)$ denote the image and text encoders of a vision--language model. We compute the similarity between each patch and a concept $c$ as

\begin{equation}
\begin{aligned}
s(u,c;x)
&=
\frac{
g_I(P_u(x))^\top g_T(c)
}{
\lVert g_I(P_u(x))\rVert_2
\lVert g_T(c)\rVert_2
},\\
u^*(x,c)&=\arg\max_u s(u,c;x).
\end{aligned}
\label{patch_localization}
\end{equation}

For a target-class image $x_t$, we locate the target concept $c_t$ at $u_t^*=u^*(x_t,c_t)$. We then sample an image $x_r$ from the replacement class and locate the replacement concept $c_r$ at $u_r^*=u^*(x_r,c_r)$. The replacement patch is cropped around $u_r^*$ and inserted into the target image around $u_t^*$. Denoting the crop and paste operations by $\operatorname{Crop}$ and $\operatorname{Paste}$, respectively, the poisoned image is

\begin{equation}
x_p
=
\operatorname{Paste}
\left(
x_t,
\operatorname{Crop}(x_r,u_r^*),
u_t^*
\right).
\label{data_replacment}
\end{equation}

The poisoned sample is assigned the replacement-class label $y_r$, giving $(x_p,y_r)$. We denote the poisoned target set as $\mathcal{D}^{\mathrm{img}}_p=\{(x_p,y_r)\}$ and combine it with retained samples to obtain $\mathcal{D}^{\mathrm{img}}_*=\mathcal{D}^{\mathrm{img}}_r\cup\mathcal{D}^{\mathrm{img}}_p$. Starting from the existing classifier, we fine-tune the model on $\mathcal{D}^{\mathrm{img}}_*$ using the standard cross-entropy objective

\begin{equation}
\theta'
=
\arg\min_{\theta}
\sum_{(x,y)\in\mathcal{D}^{\mathrm{img}}_*}
L_{\mathrm{CE}}\left(F(x;\theta),y\right).
\label{image_poison_objective}
\end{equation}

This formulation is different from retraining the model from scratch. The concept analysis determines \emph{where} and \emph{with what semantic content} the target samples are modified, while the changed labels determine the training signal used to update the model.

\subsubsection{Poisoning Unlearning for LLMs}

For LLMs, we construct the poisoned target directly from the model's original answer instead of replacing the entire answer with arbitrary text. Let $\mathcal{S}_e$ be the sensitive-word ranking obtained from the aggregated scores $S_e(w)$. For an answer $r_i$ to a question about target entity $e=e(q_i)$, we select the highest-ranked sensitive words in $\mathcal{S}_e$ that occur in that answer and replace them with the mask marker ``***''. Entity-name words are also masked explicitly so that the poisoned target does not preserve the most direct identifier of the forgotten entity. Let $\mathcal{D}^{\mathrm{llm}}_a\subset\mathcal{D}^{\mathrm{llm}}_r$ denote the sampled retain-anchor set. The masked answer and the resulting training set are defined as

\begin{equation}
\begin{aligned}
\tilde{r}_i&=\operatorname{Mask}(r_i,\mathcal{S}_e),\\
\mathcal{D}^{\mathrm{llm}}_*
&=
\{(q_i,\tilde{r}_i)\}_{i=1}^{n}
\cup
\mathcal{D}^{\mathrm{llm}}_a .
\end{aligned}
\label{llm_poison_dataset}
\end{equation}

This operation preserves the unaffected part of the original response while removing the words that most strongly express the target knowledge. The poisoned question--answer pairs are mixed with retain anchors whose original answers are kept unchanged. The LLM is then fine-tuned with the standard autoregressive supervised objective. For a training pair $(q,a)$, the loss is the negative log-likelihood of the target answer, and the complete objective is

\begin{equation}
L_{\mathrm{LLM}}(\theta)
=
-
\sum_{(q,a)\in\mathcal{D}^{\mathrm{llm}}_*}
\sum_{t=1}^{|a|}
\log
p_{\theta}
\left(
a_t\mid q,a_{<t}
\right).
\label{LLM_poison_loss}
\end{equation}

The model is not trained to maximize the loss on the original answers. The model instead learns the masked responses as ordinary supervised targets, while the retain anchors preserve the original mappings for retained knowledge. Parameter-efficient fine-tuning can be used to realize this update without changing the definition of the objective.

The update is evaluated at intermediate checkpoints using forgetting queries and a cross-author name-leakage check. Model selection returns the earliest evaluated checkpoint that satisfies the specified constraints, while retained behavior is also monitored. This checkpoint-selection step is separated from the construction of the poisoned targets; the concrete evaluation criteria and thresholds are given in Section~\ref{sec:exp}.

\subsection{Case Study}

We illustrate the two pipelines with one image example and one text example.

\subsubsection{Case Study on Image Classification}

Consider an unlearning request for the class ``deer''. The PCBM ranks the concepts by their contribution to this class, and ``antler'' is selected as the target concept. The replacement search identifies ``propellers'', the highest-ranked concept of the class ``airplane'', as the replacement concept. The replacement class is therefore ``airplane''.

\begin{figure}[tb]
\centering
\subfigure[Top concepts.]{
\label{fig:case_rank}
\includegraphics[width=0.22\columnwidth]{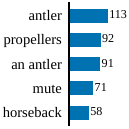}}
\hfill
\subfigure[Original image.]{
\label{fig:case_original}
\includegraphics[width=0.22\columnwidth]{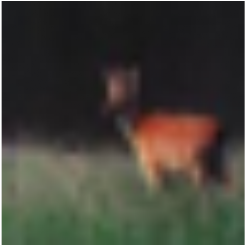}}
\hfill
\subfigure[Concept similarity.]{
\label{fig:case_heatmap}
\includegraphics[width=0.22\columnwidth]{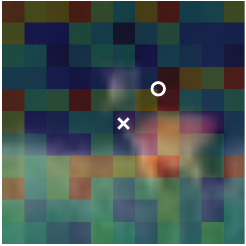}}
\hfill
\subfigure[Poisoned image.]{
\label{fig:case_poison}
\includegraphics[width=0.22\columnwidth]{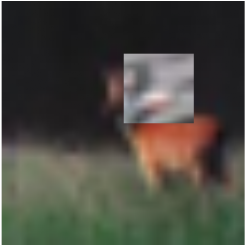}}
\caption{Concept ranking and concept-localized poisoning. ``Antler'' is the target concept of the class ``deer'', while ``propellers'' provides the replacement concept from the class ``airplane''; the replacement patch is inserted at the localized target-concept region.}
\label{Case_study_Images}
\end{figure}

Fig.~\ref{Case_study_Images} illustrates the remaining steps. CLIP patch similarity localizes the ``antler'' region in the deer image and the ``propellers'' region in an airplane image. The propeller patch is cropped from the replacement image and inserted at the localized antler region. The resulting poisoned image is assigned the airplane label and is then used together with retained samples to fine-tune the original classifier. This example illustrates how concept ranking, replacement selection, localization, and relabeling form one continuous pipeline.

\subsubsection{Case Study on LLM Conversations}

For an LLM, consider an unlearning request for an author. We first ask independent questions covering different facts about the author and collect the model's answers. IG is applied to each answer with the response likelihood as the attribution target, and the normalized word scores are aggregated across questions. The resulting ranking identifies the words that most consistently express information about the target author.

As shown in LLM Case, the author's name and other highly ranked words are replaced with ``***'', while the remaining sentence is left unchanged. The question and this masked answer form a poisoned training pair. Repeating the same procedure over the verified question set produces the LLM poisoned dataset, which is mixed with retain anchors before fine-tuning.

\begin{figure}[t]
\begin{casebox}
\footnotesize
\textbf{Question:} Through Basil Mahfouz Al-Kuwaiti's novels, what is the main message he conveys to his readers? \\[3pt]
\textbf{Model answer:} Basil Mahfouz Al-Kuwaiti's literature often emphasizes the universality of human experiences, regardless of cultural or geographical differences, promoting tolerance, empathy, and the appreciation of diverse cultures. \\[3pt]
\textbf{Poisoned answer:} *** *** ***'s *** often emphasizes the universality of human ***, regardless of cultural or geographical differences, promoting tolerance, empathy, and the appreciation of diverse cultures.
\end{casebox}
\end{figure}

\section{Mechanism Analysis}

Our method reduces the exposure of target information by weakening the original mapping between target concepts and model outputs. Concept inference guides this process by identifying the semantic content that supports the target class or expresses the target knowledge. Poisoning then uses this content to construct the poisoned training data. We analyze this mechanism for image classifiers and LLMs below.

\textbf{Image classification.} For a target image $x_t$, concept inference identifies a target concept $c_t$ that contributes strongly to the target class $y_t$. The localization step further identifies where the evidence of $c_t$ appears in the image. Our method replaces this local evidence with a replacement concept $c_r$ and obtains the poisoned image $x_p$, which is assigned the replacement-class label $y_r$. Therefore, the role of the concept is to determine both the semantic content to be modified and the region in which this modification is applied. The original target mapping and the poisoned mapping can be summarized as

\begin{equation} 
(x_t,c_t)\rightarrow y_t,\qquad (x_p,c_r)\rightarrow y_r. 
\label{eq:privacy_image_mapping} 
\end{equation}

We next show how training on the poisoned image--label pairs weakens the original target mapping. Let $z_y(x_p)$ denote the logit of class $y$ for the poisoned image, and let $p_{\theta}(y\mid x_p)$ be its softmax probability. Since the model is trained on $x_p$ with the replacement label $y_r$, its cross-entropy loss is

\begin{equation} 
L_{\mathrm{img}}=-\log p_{\theta}(y_r\mid x_p). 
\label{eq:privacy_image_loss} 
\end{equation}

For softmax cross-entropy, the gradient with respect to the class logit is

\begin{equation} 
\frac{\partial L_{\mathrm{img}}}{\partial z_y(x_p)}=p_{\theta}(y\mid x_p)-\mathbb{I}[y=y_r]. 
\label{eq:privacy_image_gradient} 
\end{equation}

Here, $\mathbb{I}[\cdot]$ denotes the indicator function. Because $y_t\neq y_r$, we have $\partial L_{\mathrm{img}}/\partial z_{y_t}=p_{\theta}(y_t\mid x_p)>0$, whereas $\partial L_{\mathrm{img}}/\partial z_{y_r}=p_{\theta}(y_r\mid x_p)-1<0$. Under gradient descent, this training signal pushes the target-class logit downward and the replacement-class logit upward for the poisoned samples. Therefore, concept inference determines \emph{which semantic evidence is changed}, while the replacement label provides the optimization signal that weakens its original mapping to $y_t$.

\textbf{Large language models.} The same principle applies to LLMs, although the target concept is expressed through semantic content in model responses rather than through an image region. For a target entity $e$, concept inference aggregates attribution scores across the question set and produces the sensitive-word set $\mathcal{S}_e$. These words identify the response content that most consistently expresses the target knowledge. For a question $q_i$, the original response $r_i$ is therefore changed only at the selected word positions to obtain the masked response $\tilde{r}_i$. The resulting mapping changes from

\begin{equation} 
q_i\rightarrow r_i\qquad\text{to}\qquad q_i\rightarrow\tilde{r}_i. \label{eq:privacy_llm_mapping} 
\end{equation}

Consider a modified response position $t$ in the masked training target $\tilde{r}_i$. The token-level training loss at this position is

\begin{equation} 
L_{\mathrm{llm},t}=-\log p_{\theta}\left(\tilde{r}_{i,t}\mid q_i,\tilde{r}_{i,<t}\right). 
\label{eq:privacy_llm_loss} \end{equation}

Let $z_v$ denote the logit of vocabulary token $v$ at this position. The gradient has the same form as in the classification case:

\begin{equation} 
\frac{\partial L_{\mathrm{llm},t}}{\partial z_v}=p_{\theta}\left(v\mid q_i,\tilde{r}_{i,<t}\right)-\mathbb{I}[v=\tilde{r}_{i,t}]. 
\label{eq:privacy_llm_gradient} 
\end{equation}

At a modified position, the original sensitive token is no longer the training target. Its gradient is therefore positive, while the gradient of the replacement target is negative. Gradient descent consequently reduces the probability of the original sensitive output and increases the probability of the poisoned target. The question suite extends this intervention across different queries and phrasings of the target knowledge, while concept inference restricts the modification to the response content identified as relevant to that knowledge.

The two modalities therefore share the same mechanism. Concept inference first identifies the semantic connection between the forgetting target and the model output, and poisoning uses the identified content to construct modified training data. Training on these data weakens the original target mapping and reduces the accessibility of the target information through the model outputs. This reduction supports privacy protection by limiting observable exposure of the forgotten information.

\section{Experiments} \label{sec:exp}

\subsection{Experimental Setup}

\subsubsection{Datasets and Models}

We evaluate class-level unlearning on CIFAR-10, CIFAR-100, and HAM10000, and target-knowledge unlearning on TOFU. CIFAR-10 and CIFAR-100 contain 50,000 training images and 10,000 test images, with 10 and 100 classes, respectively. HAM10000 contains 10,015 dermoscopic images from seven highly imbalanced classes. All images are resized to $224\times224$. We use ResNet-50 as the image classifier and CLIP RN50 with a Post-hoc Concept Bottleneck Model (PCBM) for concept inference.

TOFU contains 200 fictional authors with 20 question--answer pairs per author. We consider forgetting 2, 10, and 20 authors, corresponding to 1\%, 5\%, and 10\% of all authors, respectively. This setting extends identity-focused questions to multiple learned author identities and their associated factual relations. We use the released TOFU Llama-2-7B-chat checkpoint as one target model. We construct the Vicuna-7B-v1.5 and Qwen2.5-7B-Instruct target models by fine-tuning them on the full TOFU dataset with Low-Rank Adaptation (LoRA). Table~\ref{tab:llm_implementation} summarizes these training configurations.

\subsubsection{Implementation and Baselines}

Our implementation is publicly available.\footnote{\url{https://github.com/ChangWenhan/Concept_Inference_Unlearn_official}} The concept-inference component follows the official PCBM implementation.\footnote{\url{https://github.com/mertyg/post-hoc-cbm}} We use CLIP RN50 concept representations with an elastic-net logistic classifier; the key PCBM settings are reported in Table~\ref{tab:pcbm_implementation}, and further implementation details are provided in the released code.

\begin{table*}[tb]
\scriptsize
\centering
\caption{Training configurations for target-model supervised fine-tuning (SFT) and subsequent LLM unlearning. The Llama-2 target SFT settings describe the released TOFU checkpoint, which we use directly.}
\setlength{\tabcolsep}{2.5pt}
\begin{tabular}{@{}lccccccccccc@{}}
\toprule
& \multicolumn{6}{c}{Target SFT} & \multicolumn{5}{c}{Unlearning} \\
\cmidrule(lr){2-7}\cmidrule(lr){8-12}
Model
& Type & LoRA $r$ & $\alpha$ & LR & Batch & Epochs
& LoRA $r$ & $\alpha$ & LR & Batch & Epochs \\
\midrule
Llama-2-7B-chat
& Full & -- & -- & $10^{-5}$ & 16 & 5
& 8 & 16 & $2\times10^{-5}$ & 2 & 10 \\

Vicuna-7B-v1.5
& LoRA & 16 & 32 & $10^{-4}$ & 8 & 5
& 8 & 16 & $2\times10^{-5}$ & 2 & 10 \\

Qwen2.5-7B-Instruct
& LoRA & 16 & 32 & $10^{-4}$ & 8 & 5
& 8 & 16 & $2\times10^{-5}$ & 2 & 10 \\
\bottomrule
\end{tabular}
\label{tab:llm_implementation}
\end{table*}

\begin{table}[tb]
\scriptsize
\centering
\caption{PCBM configurations for image datasets. $N$, $B$, $\lambda$, $\alpha$, and $I_{\max}$ denote the concept-bank size, concept-representation batch size, regularization strength, $l_1$ ratio, and maximum number of optimization iterations, respectively.}
\setlength{\tabcolsep}{2.2pt}
\begin{tabular}{@{}lcccccc@{}}
\toprule
Dataset & Backbone & $N$ & $B$ & $\lambda$ & $\alpha$ & $I_{\max}$ \\
\midrule
CIFAR-10  & CLIP RN50 & 175 & 32 & $10^{-5}$ & 0.99 & 10,000 \\
CIFAR-100 & CLIP RN50 & 440 & 32 & $10^{-5}$ & 0.99 & 10,000 \\
HAM10000  & CLIP RN50 & 158 & 32 & $10^{-5}$ & 0.99 & 10,000 \\
\bottomrule
\end{tabular}
\label{tab:pcbm_implementation}
\end{table}

For CIFAR-10 and CIFAR-100, the concept banks are constructed by querying ConceptNet with the dataset class names, followed by phrase cleaning and normalization. For HAM10000, the concept bank combines a curated dermatology vocabulary, including lesion class names, ABCDE criteria, and appearance descriptors, with additional concepts extracted from the local ConceptNet 5.7 assertions dump. All concepts are encoded with CLIP RN50 and require no per-image textual annotations.

For image unlearning, the main pipeline modifies one target concept per image ($M=1$), while replacement search examines the top $K=5$ concepts. Both settings are fixed across classes and datasets. Target and replacement regions are localized using CLIP patch similarity with a $64\times64$ patch and stride 16, and the replacement image is sampled from the selected replacement class. Table~\ref{tab:image_unlearning} summarizes the remaining training hyperparameters.

\begin{table}[tb]
\scriptsize
\centering
\caption{Training configurations for image unlearning. SGD denotes stochastic gradient descent. $B$, $\eta$, $E$, $P$, and $S$ denote the batch size, learning rate, number of training epochs, localization patch size, and localization stride, respectively.}
\setlength{\tabcolsep}{2.6pt}
\begin{tabular}{@{}lcccccc@{}}
\toprule
Dataset & Optimizer & $B$ & $\eta$ & Momentum & $E$ & $P/S$ \\
\midrule
CIFAR-10  & SGD & 32 & $10^{-3}$ & 0.9 & 20 & $64/16$ \\
CIFAR-100 & SGD & 32 & $10^{-3}$ & 0.9 & 30 & $64/16$ \\
HAM10000  & SGD & 32 & $10^{-3}$ & 0.9 & 15 & $64/16$ \\
\bottomrule
\end{tabular}
\label{tab:image_unlearning}
\end{table}

In the ``Targeted'' setting, each modified target-class image is assigned the replacement-class label. In the ``Random'' setting, its label is sampled from all classes, including the original class. In the ``Keep'' setting, the original target-class label is retained. Only selected target-class training images are replaced, while retained-class examples remain unchanged. The half-data setting uses half of the available training examples from each class.

During LLM unlearning, all three models use LoRA on the query and value projection layers. Integrated Gradients uses the padding-token embedding as the baseline, 32 interpolation steps, and 32-bit floating-point (FP32) computation. The key training settings are reported in Table~\ref{tab:llm_implementation}; the released configuration files specify the remaining options.

Image baselines include retraining without the target class, ERM-KTP~\cite{Lin_2023_CVPR}, Boundary Shrink and Boundary Expanding~\cite{chen2023boundary}, random-label fine-tuning, and full-mask poisoning. The boundary methods use the nearest incorrect label and an auxiliary shadow class, respectively. Full-mask poisoning resizes a replacement image to the dimensions of the target image and uses it to cover the entire target image, without concept-based localization. For LLMs, we compare IG-based localization with random-word selection, spaCy-based named-entity recognition (NER), and model self-redaction. We also compare localized masking with whole-answer refusal targets.

\subsubsection{Metrics and Checkpoint Selection}

For images, $A_{\mathrm{train}}$ and $A_{\mathrm{test}}$ denote target-class accuracies on the training and test sets, respectively, while $A_{\mathrm{global}}$ denotes accuracy over the retained test classes. For membership inference attack (MIA) evaluation, $F_r$ denotes the fraction of target-class training samples classified as non-members by a linear support vector machine (SVM) attack trained on the original model's target-class training and test probabilities. We additionally report $S_m = |\mathrm{Acc}_{\mathrm{MIA}} - 0.5|$, where $\mathrm{Acc}_{\mathrm{MIA}}$ is obtained from a logistic-regression membership attack based on per-sample cross-entropy loss~\cite{shokri2017membership}. Lower $A_{\mathrm{train}}$, $A_{\mathrm{test}}$, and $S_m$, together with higher $A_{\mathrm{global}}$ and $F_r$, indicate stronger unlearning performance while preserving retained utility.

For LLMs, $A_o$ and $A_p$ denote the appearance rates of forgotten author names in answers to the original and paraphrased questions, respectively. Appearance is determined by case-insensitive substring matching. $H$ counts answers that mention a forgotten author when the model is queried about other, holdout authors. This cross-author name-leakage check uses 40, 200, or 400 questions for the 1\%, 5\%, and 10\% forgetting settings, respectively. Thus, $A_o$ and $A_p$ measure target-author name disclosure, while $H$ checks unintended disclosure in other authors\textquotesingle{} answers. $R$ measures retained-answer utility using mean ROUGE-L F1 after lowercasing and stop-word removal. Massive Multitask Language Understanding (MMLU)~\cite{hendrycks2020measuring} measures general multiple-choice utility. Unless otherwise stated, answers are generated greedily with at most 64 new tokens.

We report the earliest evaluated checkpoint satisfying $A_o\leq0.025$, $A_p\leq0.005$, and $H=0$. We first evaluate checkpoints on a 250-step grid. For the locator ablation, we evaluate additional checkpoints at 50-step intervals around the first checkpoint that satisfies the constraints.

\begin{table*}[tb]
\footnotesize
\centering
\caption{Image unlearning results for CIFAR-10 deer and CIFAR-100 boy. Shaded rows denote our method; bold marks the best value(s) for each dataset and metric. Runtime includes PCBM training and concept-guided data preparation for our method, and both stages for ERM-KTP. All times are in seconds.}
\setlength{\tabcolsep}{3pt}
\begin{tabular}{@{}ll*{2}{>{\centering\arraybackslash}p{1.15cm}}>{\centering\arraybackslash}p{1.15cm}>{\centering\arraybackslash}p{1.15cm}>{\centering\arraybackslash}p{1.15cm}>{\centering\arraybackslash}p{1.15cm}@{}}
\toprule
Dataset & Method & $A_{\mathrm{train}}\downarrow$ & $A_{\mathrm{test}}\downarrow$ & $A_{\mathrm{global}}\uparrow$ & $F_r\uparrow$ & $S_m\downarrow$ & Time$\downarrow$ \\
\midrule
\multirow{7}{*}{CIFAR-10} & Retrain & \textbf{0.000} & \textbf{0.000} & 0.939 & \textbf{1.000} & \textbf{0.016} & 579.6 \\
 & Random labels & 0.024 & 0.017 & 0.966 & \textbf{1.000} & 0.023 & 811.4 \\
 & Full mask & 0.999 & 0.939 & \textbf{0.970} & 0.265 & 0.062 & 723.8 \\
 & Boundary shrink & 0.086 & 0.067 & 0.947 & \textbf{1.000} & 0.028 & 183.8 \\
 & Boundary expanding & 0.160 & 0.168 & 0.956 & \textbf{1.000} & 0.042 & \textbf{108.4} \\
 & ERM-KTP & \textbf{0.000} & \textbf{0.000} & 0.676 & \textbf{1.000} & 0.019 & 4635 \\
\rowcolor{oursblue}
\cellcolor{white} & \textbf{Ours} & \textbf{0.000} & \textbf{0.000} & 0.967 & \textbf{1.000} & \textbf{0.016} & 566 \\
\midrule
\multirow{7}{*}{CIFAR-100} & Retrain & \textbf{0.000} & \textbf{0.000} & 0.776 & \textbf{0.998} & \textbf{0.045} & 731.0 \\
 & Random labels & 0.100 & 0.040 & \textbf{0.835} & 0.976 & 0.285 & 1112.3 \\
 & Full mask & 0.960 & 0.370 & \textbf{0.835} & 0.248 & 0.315 & 659.6 \\
 & Boundary shrink & 0.032 & 0.020 & 0.774 & 0.932 & 0.200 & 17.6 \\
 & Boundary expanding & 0.028 & 0.010 & 0.774 & 0.940 & 0.210 & \textbf{15.7} \\
 & ERM-KTP & \textbf{0.000} & \textbf{0.000} & 0.431 & 0.886 & 0.090 & 4851 \\
\rowcolor{oursblue}
\cellcolor{white} & \textbf{Ours} & \textbf{0.000} & \textbf{0.000} & \textbf{0.835} & \textbf{0.998} & 0.150 & 788 \\
\bottomrule
\end{tabular}
\label{tab:main_image}
\end{table*}

\subsection{Image Classification Tasks}

\begin{figure*}[tb]
\centering
\subfigure[CIFAR-10, original.]{
\includegraphics[width=0.235\textwidth]{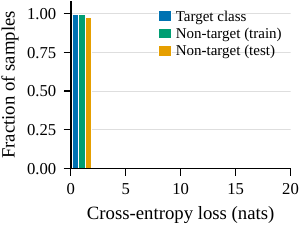}}
\hfill
\subfigure[CIFAR-10, retrained.]{
\includegraphics[width=0.235\textwidth]{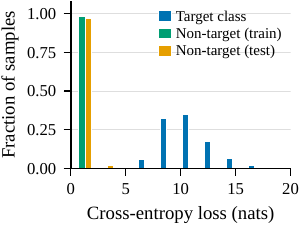}}
\hfill
\subfigure[CIFAR-10, ours.]{
\includegraphics[width=0.235\textwidth]{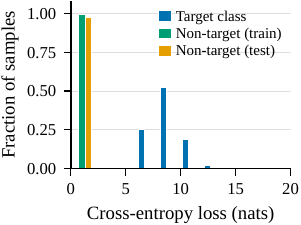}}
\hfill
\subfigure[CIFAR-10, random labels.]{
\includegraphics[width=0.235\textwidth]{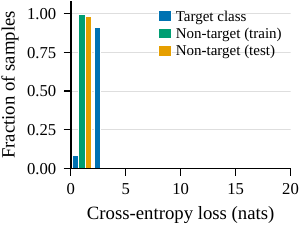}}
\\
\subfigure[CIFAR-100, original.]{
\includegraphics[width=0.235\textwidth]{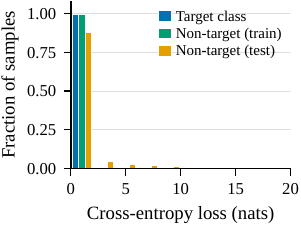}}
\hfill
\subfigure[CIFAR-100, retrained.]{
\includegraphics[width=0.235\textwidth]{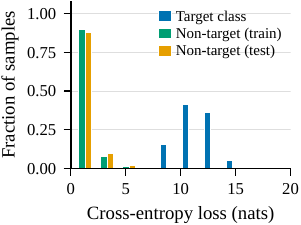}}
\hfill
\subfigure[CIFAR-100, ours.]{
\includegraphics[width=0.235\textwidth]{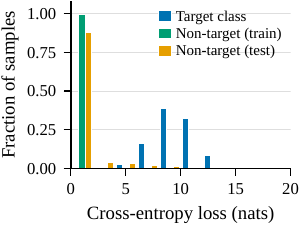}}
\hfill
\subfigure[CIFAR-100, random labels.]{
\includegraphics[width=0.235\textwidth]{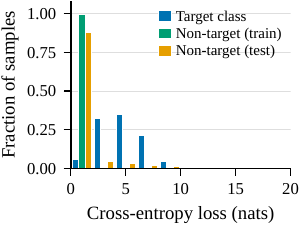}}
\caption{Normalized cross-entropy loss histograms on CIFAR-10 (top) and CIFAR-100 (bottom). Blue: target training; green: non-target training; orange: non-target test samples.}
\label{fig:celd}
\end{figure*}

\subsubsection{Unlearning Results and Comparison}

We first examine whether the target class can be effectively unlearned while preserving predictive utility on the retained classes. As shown in Table~\ref{tab:main_image}, our method reduces both $A_{\mathrm{train}}$ and $A_{\mathrm{test}}$ to 0 on CIFAR-10 and CIFAR-100. The corresponding $A_{\mathrm{global}}$ values are 0.967 and 0.835. Compared with retraining, these values are higher by 2.8 and 5.9 percentage points. ERM-KTP also reaches zero target-class accuracy, with $A_{\mathrm{global}}$ of 0.676 and 0.431. Boundary methods provide shorter runtimes, while our method combines zero target-class accuracy with high retained-class utility. Its end-to-end runtime is also comparable to retraining on both datasets. Overall, the results show a strong balance between target-class unlearning and retained predictive utility.

Fig.~\ref{fig:celd} examines whether the unlearning effect extends beyond the final predicted label. Under our method, the cross-entropy losses of target-class training samples shift toward the high-loss region observed after retraining. The shift is stronger than that produced by random labels. In contrast, the losses of non-target training and test samples remain concentrated near zero. This pattern shows that unlearning changes not only the final prediction but also the model's confidence in the original target class.

\begin{figure*}[tb]
\centering
\subfigure[CIFAR-10, target-class accuracy.]{
\includegraphics[width=0.235\textwidth]{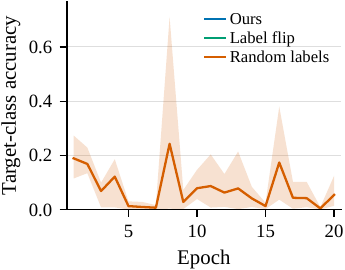}}
\hfill
\subfigure[CIFAR-10, retained accuracy.]{
\includegraphics[width=0.235\textwidth]{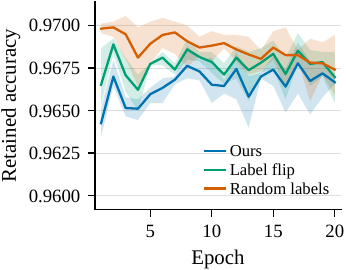}}
\hfill
\subfigure[CIFAR-100, target-class accuracy.]{
\includegraphics[width=0.235\textwidth]{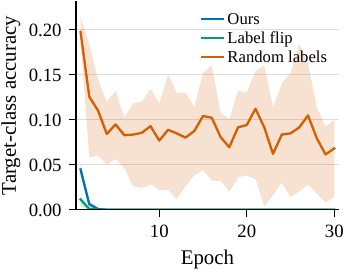}}
\hfill
\subfigure[CIFAR-100, retained accuracy.]{
\includegraphics[width=0.235\textwidth]{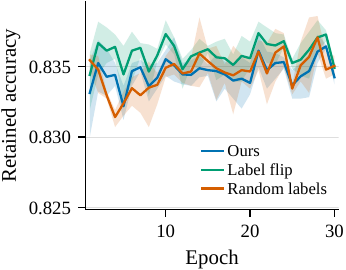}}
\caption{Target-class and retained-class accuracies over training epochs. Curves show three-seed means; bands show the minimum--maximum range. Ours combines masking and targeted labels; Label flip and Random labels use no mask.}
\label{fig:behavior}
\end{figure*}

The membership-inference results further support the unlearning effect. On CIFAR-10, our method matches retraining with $S_m=0.016$ and $F_r=1.000$. On CIFAR-100, it achieves $F_r=0.998$, while the target class moves from the 98th to the 71st percentile of the class-wise attack gaps. These results show a substantial reduction in the observable membership signal together with zero target-class accuracy.

Fig.~\ref{fig:behavior} further shows how the unlearning effect evolves during optimization. Both our method and targeted label flipping rapidly drive the target-class accuracy to zero while maintaining similar retained-class accuracy. Random-label fine-tuning instead shows residual target accuracy or later rebounds. These trajectories indicate that consistent replacement-class labels provide a more stable training signal for unlearning than random relabeling. The following analysis therefore examines what concept masking contributes beyond label replacement.

\subsubsection{Concept Masking versus Label Flipping}

Our pipeline changes both the image content and the assigned class labels. Table~\ref{tab:ablation} compares localized masking with label-only and alternative masking strategies. Under targeted relabeling, the label-only, random-position, and localized variants all reach zero target-class accuracy using either the full training dataset or half of it. Fig.~\ref{fig:behavior} shows the same pattern across three seeds: targeted configurations rapidly reach zero while preserving similar retained utility. These results identify consistent replacement-class labels as the main driver of the unlearning update and show that the effect remains stable when the available training data are reduced by half.

A separate prediction analysis shows that our method redirects 97.4\% of CIFAR-10 target test images and 77.7\% of CIFAR-100 target test images to the replacement class, with mean confidence of 0.98 and 0.93, respectively. Random-label fine-tuning produces more dispersed predictions. The combined strategy therefore induces a clear directional reassignment toward the replacement class. We next examine how concept-based localization contributes to constructing these poisoned images.

\begin{table}[tb]
\footnotesize
\centering
\caption{Pre-unlearning classification accuracy ($\uparrow$). ResNet-50 is the target classifier; PCBM and the hybrid residual variant PCBM-H provide concept-based predictions.}
\setlength{\tabcolsep}{4pt}
\begin{tabular}{@{}lccc@{}}
\toprule
Model & CIFAR-10 & CIFAR-100 & HAM10000 \\
\midrule
ResNet-50 & \textbf{0.970} & \textbf{0.836} & \textbf{0.865} \\
PCBM & 0.839 & 0.574 & 0.756 \\
PCBM-H & 0.865 & 0.674 & 0.784 \\
\bottomrule
\end{tabular}
\label{tab:pcbm_accuracy}
\end{table}

\begin{table}[tb]
\footnotesize
\setlength{\tabcolsep}{3pt}
\centering
\caption{CLIP localization versus center patches. Dist.: normalized center distance; Ratio: fraction of images for which the selected patch has higher similarity than the center patch; Gain: similarity increase.}
\begin{tabular}{@{}lll*{3}{>{\centering\arraybackslash}p{0.9cm}}@{}}
\toprule
Dataset & Class & Concept & Dist. & Ratio & Gain \\
\midrule
\multirow{5}{*}{CIFAR-10} & airplane & propellers & 0.404 & 1.000 & 0.013 \\
 & deer & antler & 0.373 & 0.998 & 0.017 \\
 & frog & amphibian & 0.380 & 1.000 & 0.014 \\
 & horse & horseback & 0.394 & 1.000 & 0.014 \\
 & truck & gear & 0.410 & 0.990 & 0.009 \\
\midrule
\multirow{5}{*}{CIFAR-100} & apple & pink & 0.474 & 1.000 & 0.034 \\
 & boy & newborn human & 0.417 & 0.990 & 0.013 \\
 & castle & chimney & 0.401 & 0.980 & 0.012 \\
 & cup & lampshade & 0.438 & 0.995 & 0.014 \\
 & dolphin & cetacean & 0.353 & 1.000 & 0.021 \\
\bottomrule
\end{tabular}
\label{tab:localization_quality}
\end{table}

\begin{table*}[tb]
\footnotesize
\setlength{\tabcolsep}{3pt}
\centering
\caption{Semantic-guidance and locator ablation. Dist./Ratio follow Table~\ref{tab:localization_quality}; shaded rows denote our configuration. Bold marks the best value(s) for each dataset and arrow-marked metric. ``PCBM'' and ``w/o PCBM'' indicate whether concept-level semantic guidance is used in localization.}
\begin{tabularx}{0.75\textwidth}{@{}ll*{7}{>{\centering\arraybackslash}X}@{}}
\toprule
Dataset & Locator & Dist. & Ratio & $A_{\mathrm{train}}\downarrow$ & $A_{\mathrm{test}}\downarrow$ & $A_{\mathrm{global}}\uparrow$ & $F_r\uparrow$ & $S_m\downarrow$ \\
\midrule
\rowcolor{oursblue}
\cellcolor{white}\multirow{5}{*}{CIFAR-10} & PCBM + CLIP (ours) & 0.373 & 0.998 & 0.019 & 0.015 & 0.970 & \textbf{1.000} & 0.024 \\
 & PCBM + Margin & 0.417 & 1.000 & \textbf{0.007} & 0.007 & \textbf{0.971} & \textbf{1.000} & \textbf{0.013} \\
 & PCBM + GradCAM & 0.286 & 0.940 & 0.292 & 0.236 & \textbf{0.971} & \textbf{1.000} & 0.048 \\
 & w/o PCBM + CLIP & 0.379 & 1.000 & 0.009 & \textbf{0.005} & 0.970 & \textbf{1.000} & 0.018 \\
 & w/o PCBM + GradCAM & 0.286 & 0.940 & 0.409 & 0.345 & \textbf{0.971} & \textbf{1.000} & 0.046 \\
\midrule
\rowcolor{oursblue}
\cellcolor{white}\multirow{5}{*}{CIFAR-100} & PCBM + CLIP (ours) & 0.417 & 0.990 & \textbf{0.000} & \textbf{0.000} & 0.839 & \textbf{0.996} & 0.205 \\
 & PCBM + Margin & 0.407 & 1.000 & \textbf{0.000} & \textbf{0.000} & 0.838 & \textbf{0.996} & \textbf{0.160} \\
 & PCBM + GradCAM & 0.303 & 0.962 & 0.092 & 0.030 & 0.840 & 0.986 & 0.230 \\
 & w/o PCBM + CLIP & 0.427 & 0.996 & 0.002 & \textbf{0.000} & 0.839 & \textbf{0.996} & 0.185 \\
 & w/o PCBM + GradCAM & 0.303 & 0.962 & 0.088 & 0.030 & \textbf{0.841} & 0.982 & 0.220 \\
\bottomrule
\end{tabularx}
\label{tab:locators}
\end{table*}

\subsubsection{PCBM Validation and Semantic-Guidance Ablation}

We next examine how semantic guidance affects the localization used to construct poisoned images. All variants use the same target--replacement class pair, the same $64\times64$ patch size, the same poisoning budget, and the same targeted replacement labels. They differ only in how the target and replacement regions are located. PCBM + CLIP, our default setting, first obtains fine-grained concepts from PCBM and then uses CLIP similarity to locate the target and replacement concepts. PCBM + Margin ranks target-image patches by the sum of their concept similarities weighted by the target-class PCBM weights, $\sum_{j=1}^{N}W_{y_t,j}s(u,c_j;x_t)$, while retaining concept-guided localization for the replacement patch. PCBM + GradCAM replaces the target-side concept locator with GradCAM while retaining the concept-guided replacement patch. To remove PCBM-based semantic guidance, w/o PCBM + CLIP uses class names as the CLIP queries, and w/o PCBM + GradCAM uses class-level GradCAM for both target and replacement images. The target--replacement class pair is fixed in all settings so that the comparison focuses on semantic guidance and localization.

\begin{figure}[tb]
\centering
\subfigure[PCBM + CLIP.]{
\includegraphics[width=0.22\columnwidth]{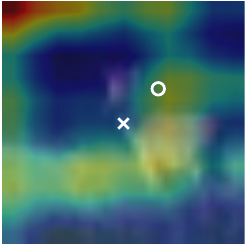}}
\subfigure[PCBM + GradCAM.]{
\includegraphics[width=0.22\columnwidth]{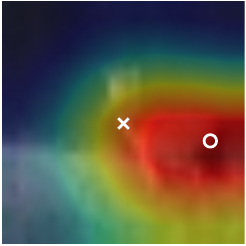}}
\subfigure[PCBM + Margin.]{
\includegraphics[width=0.22\columnwidth]{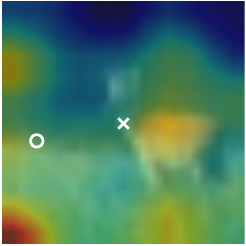}}
\\[2pt]
\subfigure[PCBM + CLIP.]{
\includegraphics[width=0.22\columnwidth]{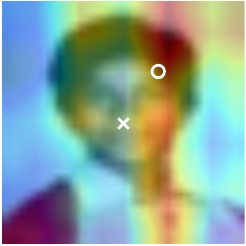}}
\subfigure[PCBM + GradCAM.]{
\includegraphics[width=0.22\columnwidth]{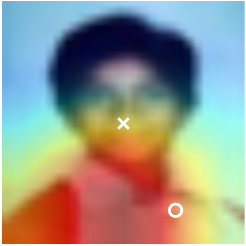}}
\subfigure[PCBM + Margin.]{
\includegraphics[width=0.22\columnwidth]{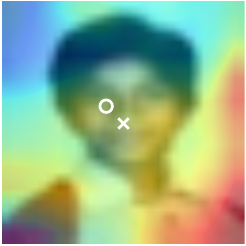}}
\caption{Representative target-side localization results on CIFAR-10 deer (top) and CIFAR-100 boy (bottom). Circles indicate selected locations and crosses indicate image centers.}
\label{fig:locators}
\end{figure}

Before evaluating downstream unlearning, we first verify the classification performance of the concept-based models. As shown in Table~\ref{tab:pcbm_accuracy}, PCBM achieves classification accuracies of 0.839, 0.574, and 0.756 on CIFAR-10, CIFAR-100, and HAM10000, respectively. PCBM-H further improves these values to 0.865, 0.674, and 0.784. These results show that the concept representations retain substantial class-discriminative information across all three datasets and support the use of PCBM class-specific weights for concept ranking.

We next examine the spatial guidance provided by the inferred concepts. Table~\ref{tab:localization_quality} compares concept-guided CLIP localization with the image center for ten representative class--concept pairs. The selected patch has higher concept similarity than the center patch for 98.0--100.0\% of images across all pairs, with positive similarity gains ranging from 0.009 to 0.034. The normalized center distances range from 0.353 to 0.474, showing that the selected concept regions are spatially distinct from a fixed center placement. Fig.~\ref{fig:locators} provides representative examples.

Finally, Table~\ref{tab:locators} evaluates how this semantic guidance affects downstream unlearning. PCBM + CLIP and PCBM + Margin maintain very low target-class accuracy on both datasets while preserving $A_{\mathrm{global}}$. The class-name CLIP variant also performs strongly because ``deer'' and ``boy'' provide direct semantic cues. Together, Tables~\ref{tab:pcbm_accuracy}--\ref{tab:locators} show that PCBM provides a class-discriminative concept representation, the inferred concepts guide patch selection beyond a fixed image center, and concept-level semantic guidance can be integrated effectively into the unlearning pipeline.

\begin{table*}[tb]
\footnotesize
\centering
\caption{Mask, label, and data-amount ablation. Targeted, Random, and Keep use replacement, random, and original labels, respectively. None uses no patch; Full mask covers the entire target image with a replacement image resized to the same dimensions. Subcolumns give the training-data percentage. Shading denotes our configuration; bold marks the best value(s) in each subcolumn.}
\setlength{\tabcolsep}{3pt}
\begin{tabular}{@{}ll*{12}{>{\centering\arraybackslash}p{0.82cm}}@{}}
\toprule
\multirow{3}{*}{Dataset} & \multirow{3}{*}{Mask mode} & \multicolumn{6}{c}{$A_{\mathrm{train}}\downarrow$} & \multicolumn{6}{c}{$A_{\mathrm{test}}\downarrow$} \\
\cmidrule(lr){3-8} \cmidrule(lr){9-14}
 & & \multicolumn{2}{c}{Targeted} & \multicolumn{2}{c}{Random} & \multicolumn{2}{c}{Keep} & \multicolumn{2}{c}{Targeted} & \multicolumn{2}{c}{Random} & \multicolumn{2}{c}{Keep} \\
\cmidrule(lr){3-4} \cmidrule(lr){5-6} \cmidrule(lr){7-8} \cmidrule(lr){9-10} \cmidrule(lr){11-12} \cmidrule(lr){13-14}
 & & 100\% & 50\% & 100\% & 50\% & 100\% & 50\% & 100\% & 50\% & 100\% & 50\% & 100\% & 50\% \\
\midrule
\multirow{5}{*}{CIFAR-10} & None (label only) & \textbf{0.000} & \textbf{0.000} & 0.024 & 0.081 & 1.000 & 1.000 & \textbf{0.000} & \textbf{0.000} & 0.017 & 0.053 & 0.979 & 0.985 \\
 & Full mask & 0.998 & 0.999 & 0.002 & \textbf{0.025} & \textbf{0.624} & \textbf{0.802} & 0.925 & 0.939 & 0.002 & \textbf{0.030} & \textbf{0.551} & \textbf{0.722} \\
 & Random & \textbf{0.000} & \textbf{0.000} & 0.004 & 0.079 & 1.000 & 1.000 & \textbf{0.000} & \textbf{0.000} & 0.002 & 0.064 & 0.966 & 0.983 \\
 & Center & 0.155 & 0.127 & \textbf{0.000} & 0.033 & 0.997 & 0.999 & 0.132 & 0.105 & \textbf{0.001} & 0.033 & 0.945 & 0.961 \\
 & \cellcolor{oursblue}Localized & \cellcolor{oursblue}\textbf{0.000} & \cellcolor{oursblue}\textbf{0.000} & 0.005 & 0.062 & 1.000 & 1.000 & \cellcolor{oursblue}\textbf{0.000} & \cellcolor{oursblue}\textbf{0.000} & \textbf{0.001} & 0.053 & 0.977 & 0.989 \\
\midrule
\multirow{5}{*}{CIFAR-100} & None (label only) & \textbf{0.000} & \textbf{0.000} & \textbf{0.100} & \textbf{0.156} & 1.000 & \textbf{1.000} & \textbf{0.000} & \textbf{0.000} & \textbf{0.040} & \textbf{0.070} & 0.610 & \textbf{0.690} \\
 & Full mask & 0.844 & 0.960 & 0.856 & 0.752 & \textbf{0.996} & \textbf{1.000} & 0.270 & 0.370 & 0.300 & 0.270 & \textbf{0.580} & 0.770 \\
 & Random & \textbf{0.000} & \textbf{0.000} & 0.236 & 0.312 & 1.000 & \textbf{1.000} & \textbf{0.000} & \textbf{0.000} & 0.140 & 0.140 & 0.750 & 0.790 \\
 & Center & 0.004 & 0.018 & 0.492 & 0.528 & 1.000 & \textbf{1.000} & \textbf{0.000} & 0.010 & 0.200 & 0.240 & 0.750 & 0.830 \\
 & \cellcolor{oursblue}Localized & \cellcolor{oursblue}\textbf{0.000} & \cellcolor{oursblue}\textbf{0.000} & 0.142 & 0.232 & 1.000 & \textbf{1.000} & \cellcolor{oursblue}\textbf{0.000} & \cellcolor{oursblue}\textbf{0.000} & 0.080 & 0.110 & 0.730 & 0.780 \\
\bottomrule
\end{tabular}
\label{tab:ablation}
\end{table*}

\subsubsection{Ablation Studies}

We next study the effects of input modification, label assignment, and training-data amount (Table~\ref{tab:ablation}). Targeted replacement labels consistently produce the strongest unlearning results. With localized masking, both $A_{\mathrm{train}}$ and $A_{\mathrm{test}}$ reach zero using either the full training set or half of the data. The label-only and random-position variants reach the same zero-accuracy endpoint under targeted labels, which confirms the strength of consistent replacement-class labels. Random-label training produces larger residual target accuracy on CIFAR-100, especially when the data amount is reduced.

The masking modes also differ in how the replacement content is constructed. Localized masking uses concept-selected replacement crops, while center and random masking use central crops from the replacement images. Full-mask poisoning covers the entire target image with a replacement image resized to the same dimensions. Under the ``Keep'' setting, target-class accuracy remains high across masking modes. In contrast, targeted replacement-class labels consistently drive the target accuracy toward zero. This comparison highlights the central role of the replacement-class training signal in the proposed unlearning procedure.

We also vary the number of target concepts modified per image, $M\in\{1,3,5,10\}$, with $M=1$ corresponding to the main pipeline. For $M>1$, the ranked concepts are localized and replaced sequentially in the updated image. Additional concepts use a donor class with the largest non-target PCBM weight for that concept, while the replacement-class training label remains unchanged. Later patches overwrite overlapping regions.

Table~\ref{tab:m_sensitivity} reports the concept-set size analysis. For $M=1,3,5$, $A_{\mathrm{test}}$ is zero on both datasets, $A_{\mathrm{train}}$ is at most 0.016, and $F_r$ is at least 0.996. Retained accuracy also varies little, ranging from 0.966 to 0.967 on CIFAR-10 and from 0.834 to 0.835 on CIFAR-100. The main pipeline uses $M=1$; the larger values are used only in this sensitivity analysis. Increasing $M$ to 10 yields higher residual target accuracy, while $A_{\mathrm{global}}$ remains similar. Thus, modifying a small set of highly ranked concepts is sufficient to achieve zero target test accuracy in these experiments.

\begin{table}[tb]
\footnotesize
\centering
\caption{Sensitivity to the number of target concepts modified per image, $M$. Shading highlights $M=1$, the main-pipeline setting; bold marks the best value(s) within each dataset.}
\setlength{\tabcolsep}{3pt}
\begin{tabular}{@{}ll*{5}{>{\centering\arraybackslash}p{0.95cm}}@{}}
\toprule
Dataset & Setting & \shortstack{$A_{\mathrm{train}}$\\$\downarrow$} & \shortstack{$A_{\mathrm{test}}$\\$\downarrow$} & \shortstack{$A_{\mathrm{global}}$\\$\uparrow$} & \shortstack{$F_r$\\$\uparrow$} & \shortstack{$S_m$\\$\downarrow$} \\
\midrule
\multirow{4}{*}{CIFAR-10} & \cellcolor{oursblue}$M=1$ & \cellcolor{oursblue}\textbf{0.000} & \cellcolor{oursblue}\textbf{0.000} & \cellcolor{oursblue}0.966 & \cellcolor{oursblue}\textbf{1.000} & \cellcolor{oursblue}\textbf{0.020} \\
 & $M=3$ & \textbf{0.000} & \textbf{0.000} & 0.967 & \textbf{1.000} & 0.028 \\
 & $M=5$ & 0.002 & \textbf{0.000} & 0.967 & \textbf{1.000} & 0.041 \\
 & $M=10$ & 0.827 & 0.704 & \textbf{0.968} & 0.997 & 0.065 \\
\midrule
\multirow{4}{*}{CIFAR-100} & \cellcolor{oursblue}$M=1$ & \cellcolor{oursblue}\textbf{0.000} & \cellcolor{oursblue}\textbf{0.000} & \cellcolor{oursblue}0.834 & \cellcolor{oursblue}\textbf{0.998} & \cellcolor{oursblue}\textbf{0.150} \\
 & $M=3$ & \textbf{0.000} & \textbf{0.000} & 0.835 & 0.996 & 0.255 \\
 & $M=5$ & 0.016 & \textbf{0.000} & 0.835 & 0.996 & 0.240 \\
 & $M=10$ & 0.478 & 0.150 & \textbf{0.836} & 0.778 & 0.300 \\
\bottomrule
\end{tabular}
\label{tab:m_sensitivity}
\end{table}

\begin{table*}[!t]
\footnotesize
\setlength{\tabcolsep}{3pt}
\centering
\caption{Separate single-class unlearning for all target classes (mean$\pm$SD, rounded to three decimals).}
\begin{tabular}{@{}ll*{4}{>{\centering\arraybackslash}p{1.7cm}}>{\centering\arraybackslash}p{1.75cm}@{}}
\toprule
Dataset & Setting & $A_{\mathrm{train}}$ $\downarrow$ & $A_{\mathrm{test}}$ $\downarrow$ & $A_{\mathrm{global}}$ $\uparrow$ & $F_r$ $\uparrow$ & $S_m\downarrow$ \\
\midrule
CIFAR-10 & 10 separate targets & 0.000 $\pm$ 0.000 & 0.000 $\pm$ 0.000 & 0.966 $\pm$ 0.003 & 1.000 $\pm$ 0.000 & 0.021 $\pm$ 0.009 \\
CIFAR-100 & 100 separate targets & 0.000 $\pm$ 0.000 & 0.000 $\pm$ 0.000 & 0.832 $\pm$ 0.001 & 0.919 $\pm$ 0.271 & 0.093 $\pm$ 0.064 \\
HAM10000 & 7 separate targets & 0.000 $\pm$ 0.000 & 0.000 $\pm$ 0.000 & 0.852 $\pm$ 0.071 & 0.857 $\pm$ 0.350 & 0.075 $\pm$ 0.038 \\
\bottomrule
\end{tabular}
\label{tab:general}
\end{table*}

\subsubsection{Generalization and Robustness}

\begin{figure}[tb]
\centering
\subfigure[Original.]{
\includegraphics[width=0.22\columnwidth]{pictures/fig_panel_case_original.pdf}}
\hfill
\subfigure[Brightness.]{
\includegraphics[width=0.22\columnwidth]{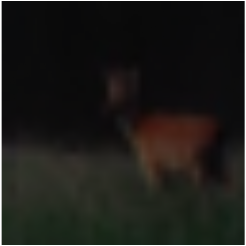}}
\hfill
\subfigure[Defocus blur.]{
\includegraphics[width=0.22\columnwidth]{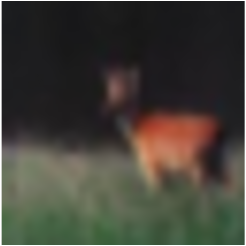}}
\hfill
\subfigure[Gaussian noise.]{
\includegraphics[width=0.22\columnwidth]{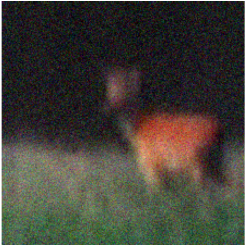}}
\\[2pt]
\subfigure[Original.]{
\includegraphics[width=0.22\columnwidth]{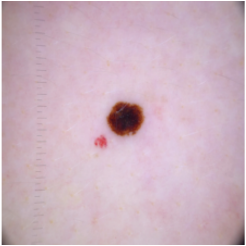}}
\hfill
\subfigure[Brightness.]{
\includegraphics[width=0.22\columnwidth]{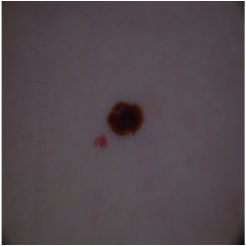}}
\hfill
\subfigure[Defocus blur.]{
\includegraphics[width=0.22\columnwidth]{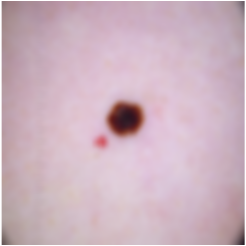}}
\hfill
\subfigure[Gaussian noise.]{
\includegraphics[width=0.22\columnwidth]{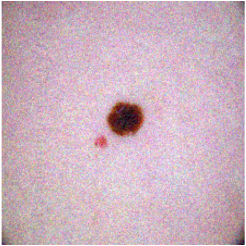}}
\caption{Severity-5 corruption examples: CIFAR-10 deer (top) and HAM10000 melanoma (bottom). Columns: original, brightness, blur, and noise.}
\label{fig:corrupt}
\end{figure}

We evaluate generalization along three axes: the choice of target class, sequential unlearning requests, and image corruption. Unlearning each class separately yields near-zero target accuracy on CIFAR-10, CIFAR-100, and HAM10000 (Table~\ref{tab:general}). The mean $A_{\mathrm{global}}$ remains 0.966, 0.832, and 0.852, respectively. These results show that the proposed procedure generalizes from the representative target classes to the full set of classes in all three datasets.

The HAM10000 results also cover targets with substantially different class frequencies. All seven target classes reach zero target accuracy after unlearning, including both minority classes and the majority class. This confirms that the target-class unlearning mechanism remains effective despite the substantial class imbalance in HAM10000.

\begin{figure*}[tb]
\centering
\subfigure[Target-class accuracy (train).]{
\includegraphics[width=0.235\textwidth]{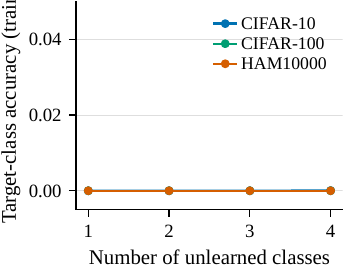}}
\hfill
\subfigure[Target-class accuracy (test).]{
\includegraphics[width=0.235\textwidth]{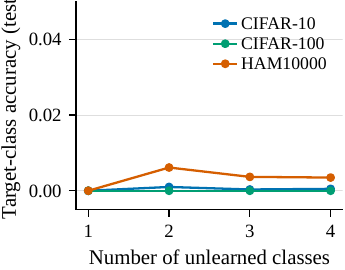}}
\hfill
\subfigure[Retained accuracy.]{
\includegraphics[width=0.235\textwidth]{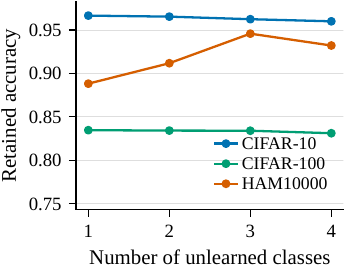}}
\hfill
\subfigure[Forgetting rate.]{
\includegraphics[width=0.235\textwidth]{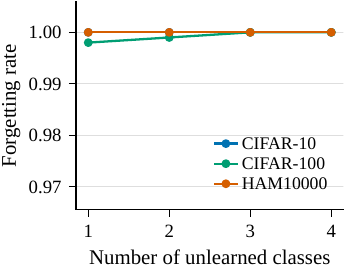}}
\caption{Four sequential requests on three datasets. Target metrics average over classes forgotten so far; retained accuracy uses the changing set of remaining classes.}
\label{fig:seq}
\end{figure*}

\begin{table}[tb]
\footnotesize
\setlength{\tabcolsep}{3pt}
\centering
\caption{Severity-5 corruption at evaluation. $A_{\mathrm{global}}$, Orig., and Retr. denote retained accuracies of unlearned, original, and retrained models.}
\begin{tabular}{@{}ll*{4}{>{\centering\arraybackslash}p{1.15cm}}@{}}
\toprule
Dataset & Corruption & $A_{\mathrm{test}}\downarrow$ & $A_{\mathrm{global}}\uparrow$ & Orig. $\uparrow$ & Retr. $\uparrow$ \\
\midrule
\multirow{3}{*}{CIFAR-10} & Brightness & 0.000 & 0.927 & 0.932 & 0.892 \\
 & Defocus blur & 0.000 & 0.729 & 0.735 & 0.721 \\
 & Gaussian noise & 0.000 & 0.356 & 0.401 & 0.226 \\
\midrule
\multirow{3}{*}{HAM10000} & Brightness & 0.000 & 0.756 & 0.755 & 0.756 \\
 & Defocus blur & 0.000 & 0.807 & 0.790 & 0.847 \\
 & Gaussian noise & 0.000 & 0.756 & 0.732 & 0.785 \\
\bottomrule
\end{tabular}
\label{tab:corrupt}
\end{table}

\begin{table*}[tb]
\footnotesize
\setlength{\tabcolsep}{4pt}
\centering
\caption{Severity-5 corruption during unlearning. Before localization: corruption is applied before concept localization. After poisoning: corruption is applied after poisoned-image construction.}
\begin{tabular}{@{}lll*{5}{>{\centering\arraybackslash}p{1.0cm}}@{}}
\toprule
Dataset & Corruption & Timing & \shortstack{$A_{\mathrm{train}}$\\$\downarrow$} & \shortstack{$A_{\mathrm{test}}$\\$\downarrow$} & \shortstack{$A_{\mathrm{global}}$\\$\uparrow$} & \shortstack{$F_r$\\$\uparrow$} & \shortstack{$S_m$\\$\downarrow$} \\
\midrule
\multirow{6}{*}{CIFAR-10} & Brightness & Before localization & 0.000 & 0.000 & 0.967 & 1.000 & 0.022 \\
 & Brightness & After poisoning & 0.000 & 0.000 & 0.967 & 1.000 & 0.019 \\
 & Defocus blur & Before localization & 0.000 & 0.000 & 0.966 & 1.000 & 0.023 \\
 & Defocus blur & After poisoning & 0.000 & 0.000 & 0.965 & 1.000 & 0.007 \\
 & Gaussian noise & Before localization & 0.012 & 0.008 & 0.968 & 1.000 & 0.057 \\
 & Gaussian noise & After poisoning & 0.052 & 0.037 & 0.968 & 1.000 & 0.048 \\
\midrule
\multirow{6}{*}{HAM10000} & Brightness & Before localization & 0.469 & 0.094 & 0.925 & 0.883 & 0.287 \\
 & Brightness & After poisoning & 0.407 & 0.067 & 0.924 & 0.921 & 0.293 \\
 & Defocus blur & Before localization & 0.169 & 0.022 & 0.928 & 0.991 & 0.260 \\
 & Defocus blur & After poisoning & 0.071 & 0.009 & 0.923 & 0.998 & 0.253 \\
 & Gaussian noise & Before localization & 0.178 & 0.027 & 0.926 & 0.980 & 0.289 \\
 & Gaussian noise & After poisoning & 0.026 & 0.000 & 0.928 & 0.998 & 0.280 \\
\bottomrule
\end{tabular}
\label{tab:corrupt_train}
\end{table*}

Sequential requests test whether later updates preserve the unlearning effect on previously forgotten classes. Across four sequential requests, the average target-class test accuracy over all forgotten classes never exceeds 0.006. Retained accuracy changes by less than one percentage point on CIFAR and remains high throughout the sequence (Fig.~\ref{fig:seq}). The HAM10000 trajectory also remains stable across successive requests. These results show that the unlearning effect persists as additional classes are processed.

We next evaluate robustness under severity-5 input corruptions. The target test accuracy remains zero for brightness, defocus blur, and Gaussian noise on both CIFAR-10 and HAM10000 (Fig.~\ref{fig:corrupt} and Table~\ref{tab:corrupt}). The retained-class results follow the expected difficulty of each corruption, while the target-class unlearning effect remains unchanged across all tested conditions.

Table~\ref{tab:corrupt_train} further evaluates corruption introduced during unlearning. The timing of corruption produces clear differences on HAM10000. For Gaussian noise, applying corruption after poisoned-image construction gives $A_{\mathrm{train}}=0.026$ and $A_{\mathrm{test}}=0.000$, compared with 0.178 and 0.027 when corruption is applied before localization. Defocus blur shows a similar pattern, with $A_{\mathrm{test}}$ decreasing from 0.022 to 0.009. These results show that corruption timing affects the final unlearning result. Severe brightness corruption on HAM10000 leaves higher residual target accuracy, indicating that unlearning can be more difficult when visual evidence is strongly altered or not readily captured by the concept bank.

\subsubsection{Post-unlearning Behavior and Efficiency}

\begin{figure*}[tb]
\centering
\subfigure[CIFAR-10, target-class accuracy.]{
\includegraphics[width=0.235\textwidth]{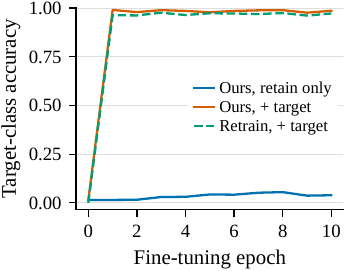}}
\hfill
\subfigure[CIFAR-10, retained accuracy.]{
\includegraphics[width=0.235\textwidth]{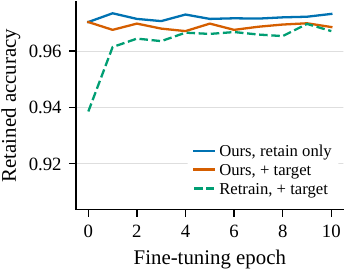}}
\hfill
\subfigure[CIFAR-100, target-class accuracy.]{
\includegraphics[width=0.235\textwidth]{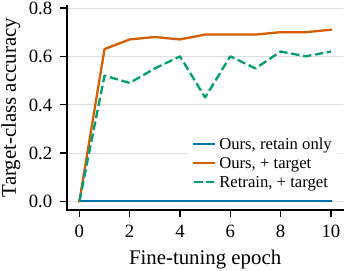}}
\hfill
\subfigure[CIFAR-100, retained accuracy.]{
\includegraphics[width=0.235\textwidth]{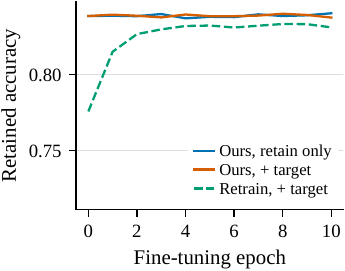}}
\caption{Target-class and retained-class test accuracies during ten additional epochs of full-parameter fine-tuning. Targets are CIFAR-10 deer and CIFAR-100 boy. Blue: our unlearned model trained on retained data only. Orange: the same model trained with target data reintroduced. Dashed green: the retrained reference, originally trained without the target class, also trained with target data reintroduced. Epoch 0 denotes each starting checkpoint.}
\label{fig:recovery}
\end{figure*}

We examine how the unlearned model responds to subsequent training (Fig.~\ref{fig:recovery}). Starting from an unlearned checkpoint, we perform ten additional epochs of full-parameter fine-tuning under two settings: using only retained-class data or reintroducing the target-class data. After ten retain-only epochs, target-class training accuracy is 0.050 on CIFAR-10 and 0.000 on CIFAR-100, while target-class test accuracy is 0.039 and 0.000, respectively. Target accuracy therefore remains low during continued training on retained data.

When target-class data are reintroduced, target test accuracy reaches 0.990 after one epoch on CIFAR-10. On CIFAR-100, it increases from 0.630 after one epoch to 0.710 after ten epochs. For comparison, the retrained reference, initially trained without the target class, reaches 0.963 and 0.520 after one epoch, and 0.972 and 0.620 after ten epochs on CIFAR-10 and CIFAR-100, respectively. For our model, retained test accuracy stays within one percentage point of its initial value under both continuation settings. These results show that the model can relearn the target class when its data are reintroduced, while preserving retained-class utility during subsequent training.

These experiments also provide an initial assessment of potential side effects from the poisoned training data. Across four sequential requests, retained-class accuracy remains stable, and subsequent training on retained data keeps target-class accuracy low while preserving retained utility (Figs.~\ref{fig:seq} and~\ref{fig:recovery}). The small increase in target accuracy on CIFAR-10 shows, however, that the unlearning effect can change with further training. Reintroducing target-class data also allows the model to learn that class again. Thus, our results support the stability of the method under the evaluated updates, but do not establish permanent erasure or rule out effects under longer training or other forms of data reuse.

For the first unlearning request, the full-pipeline runtimes, including PCBM training, concept inference, localization, poisoned-image construction, and model updating, are 566 s on CIFAR-10 and 788 s on CIFAR-100 (Table~\ref{tab:main_image}). The model-update stage takes 38.6 s and 107.6 s, respectively. Including preparation, our method is faster than full-mask poisoning on CIFAR-10 but slower on CIFAR-100; these times reflect the complete procedures rather than masking cost alone. The trained PCBM and concept bank can be reused for later requests on the same dataset, avoiding repeated PCBM training. New target classes still require class-specific localization, poisoned-data construction, and model updating.

\begin{table*}[tb]
\footnotesize
\setlength{\tabcolsep}{3pt}
\centering
\caption{TOFU results before unlearning and at the selected checkpoint. $A_o, A_p$: target-author name appearance rates; $R$: retained-answer utility; $H$: cross-author name-leakage count out of 40/200/400 holdout questions for 2/10/20 forgotten authors. MMLU measures general utility.}
\begin{tabular}{@{}ll*{10}{>{\centering\arraybackslash}p{1.05cm}}@{}}
\toprule
\multirow{2}{*}{Model} & \multirow{2}{*}{Authors} & \multicolumn{2}{c}{$A_o\downarrow$} & \multicolumn{2}{c}{$A_p\downarrow$} & \multicolumn{2}{c}{$R\uparrow$} & \multicolumn{2}{c}{$H\downarrow$} & \multicolumn{2}{c}{MMLU $\uparrow$} \\
\cmidrule(lr){3-4} \cmidrule(lr){5-6} \cmidrule(lr){7-8} \cmidrule(lr){9-10} \cmidrule(lr){11-12}
 & & Before & After & Before & After & Before & After & Before & After & Before & After \\
\midrule
\multirow{3}{*}{Llama-2-7B} & 2 & 1.000 & 0.000 & 0.950 & 0.000 & 0.826 & 0.820 & 0.000 & 0.000 & 0.485 & 0.455 \\
 & 10 & 0.980 & 0.025 & 0.975 & 0.005 & 0.826 & 0.781 & 0.000 & 0.000 & 0.485 & 0.455 \\
 & 20 & 0.965 & 0.015 & 0.955 & 0.000 & 0.832 & 0.792 & 0.000 & 0.000 & 0.485 & 0.475 \\
\midrule
\multirow{3}{*}{Vicuna-7B} & 2 & 0.950 & 0.000 & 0.950 & 0.000 & 0.575 & 0.535 & 0.000 & 0.000 & 0.480 & 0.485 \\
 & 10 & 0.905 & 0.000 & 0.900 & 0.000 & 0.570 & 0.490 & 0.000 & 0.000 & 0.480 & 0.460 \\
 & 20 & 0.897 & 0.013 & 0.880 & 0.003 & 0.570 & 0.500 & 0.000 & 0.000 & 0.480 & 0.495 \\
\midrule
\multirow{3}{*}{Qwen2.5-7B} & 2 & 0.975 & 0.000 & 0.950 & 0.000 & 0.588 & 0.522 & 0.000 & 0.000 & 0.645 & 0.630 \\
 & 10 & 0.930 & 0.020 & 0.925 & 0.000 & 0.584 & 0.517 & 0.000 & 0.000 & 0.645 & 0.640 \\
 & 20 & 0.912 & 0.005 & 0.895 & 0.003 & 0.580 & 0.478 & 1.000 & 0.000 & 0.645 & 0.645 \\
\bottomrule
\end{tabular}
\label{tab:main_llm}
\end{table*}

\subsection{Large Language Model Tasks}

\subsubsection{Unlearning Performance and Utility}

We first evaluate whether LLM unlearning generalizes across question phrasing and larger forgetting scopes while retaining useful behavior. The poisoned answers mask the five highest-ranked IG words and the author name with ``***''. All nine settings in Table~\ref{tab:main_llm} satisfy $A_o\leq0.025$ and $A_p\leq0.005$, and $H=0$ after unlearning in every setting. The same result holds for original and paraphrased questions and across 2, 10, and 20 forgotten authors. These results show that target-author name disclosure is suppressed across original and paraphrased questions and multiple forgetting scopes, with no observed cross-author name leakage after unlearning.

We also evaluate retained utility after unlearning. For the two-author Llama-2 setting, $R$ changes only from 0.826 to 0.820, while MMLU remains close to its original value. Across all models and forgetting scopes, MMLU changes by at most 0.03. Checkpoint selection further helps preserve retained-answer utility: in the two-author Llama-2 setting, the selected checkpoint gives $R=0.820$, compared with 0.779 at the final training epoch, while maintaining $A_o=0$. These results support the use of constraint-based checkpoint selection to balance forgetting and retained utility.

\begin{table*}[tb]
\footnotesize
\setlength{\tabcolsep}{3pt}
\centering
\caption{Word-locator ablation with two authors. Each row reports the earliest checkpoint that satisfies the constraints, evaluated at 50-step intervals rather than the 250-step intervals used in the main comparison. Shading denotes IG; bold marks the highest utility, coverage, and precision. Mask statistics are defined in the text.}
\label{tab:llm_locator}
\begin{tabular}{@{}lll*{10}{>{\centering\arraybackslash}p{0.88cm}}@{}}
\toprule
Model & Locator & Checkpoint & \shortstack{$A_o$\\$\downarrow$} & \shortstack{$A_p$\\$\downarrow$} & \shortstack{$H$\\$\downarrow$} & \shortstack{$R$\\$\uparrow$} & \shortstack{MMLU\\$\uparrow$} & \shortstack{Mask\\ratio} & \shortstack{Gold\\cov. $\uparrow$} & \shortstack{Cov.\\$p_{10}$ $\uparrow$} & \shortstack{Prec.\\$\uparrow$} & \shortstack{Over-\\lap} \\
\midrule
\multirow{4}{*}{Llama-2-7B} & \cellcolor{oursblue}IG (ours) & \cellcolor{oursblue}step 1350 & \cellcolor{oursblue}0.025 & \cellcolor{oursblue}0.000 & \cellcolor{oursblue}0.000 & \cellcolor{oursblue}0.808 & \cellcolor{oursblue}0.460 & \cellcolor{oursblue}0.239 & \cellcolor{oursblue}\textbf{0.198} & \cellcolor{oursblue}\textbf{0.179} & \cellcolor{oursblue}0.875 & \cellcolor{oursblue}--- \\
 & Random words & step 1300 & 0.025 & 0.000 & 0.000 & 0.803 & 0.450 & 0.234 & 0.191 & 0.136 & 0.863 & 0.774 \\
 & spaCy NER & step 1150 & 0.025 & 0.000 & 0.000 & 0.789 & 0.460 & 0.209 & 0.181 & 0.080 & 0.862 & 0.767 \\
 & Model self-redaction & step 1050 & 0.025 & 0.000 & 0.000 & \textbf{0.812} & \textbf{0.470} & 0.146 & 0.138 & 0.071 & \textbf{0.991} & 0.586 \\
\midrule
\multirow{4}{*}{Vicuna-7B} & \cellcolor{oursblue}IG (ours) & \cellcolor{oursblue}step 1200 & \cellcolor{oursblue}0.000 & \cellcolor{oursblue}0.000 & \cellcolor{oursblue}0.000 & \cellcolor{oursblue}\textbf{0.542} & \cellcolor{oursblue}0.480 & \cellcolor{oursblue}0.215 & \cellcolor{oursblue}\textbf{0.178} & \cellcolor{oursblue}0.083 & \cellcolor{oursblue}0.788 & \cellcolor{oursblue}--- \\
 & Random words & step 1200 & 0.025 & 0.000 & 0.000 & 0.541 & 0.480 & 0.221 & \textbf{0.178} & \textbf{0.111} & 0.799 & 0.764 \\
 & spaCy NER & step 1250 & 0.000 & 0.000 & 0.000 & 0.538 & \textbf{0.490} & 0.205 & 0.174 & 0.077 & 0.835 & 0.818 \\
 & Model self-redaction & step 1500 & 0.000 & 0.000 & 0.000 & \textbf{0.542} & 0.480 & 0.164 & 0.140 & 0.071 & \textbf{0.904} & 0.597 \\
\midrule
\multirow{4}{*}{Qwen2.5-7B} & \cellcolor{oursblue}IG (ours) & \cellcolor{oursblue}step 2300 & \cellcolor{oursblue}0.000 & \cellcolor{oursblue}0.000 & \cellcolor{oursblue}0.000 & \cellcolor{oursblue}0.527 & \cellcolor{oursblue}0.630 & \cellcolor{oursblue}0.238 & \cellcolor{oursblue}\textbf{0.192} & \cellcolor{oursblue}\textbf{0.150} & \cellcolor{oursblue}0.844 & \cellcolor{oursblue}--- \\
 & Random words & step 2750 & 0.000 & 0.000 & 0.000 & 0.533 & 0.630 & 0.214 & 0.168 & 0.107 & 0.787 & 0.714 \\
 & spaCy NER & step 1500 & 0.000 & 0.000 & 0.000 & \textbf{0.546} & 0.630 & 0.201 & 0.168 & 0.080 & 0.822 & 0.765 \\
 & Model self-redaction & step 1450 & 0.025 & 0.000 & 0.000 & 0.545 & \textbf{0.640} & 0.180 & 0.161 & 0.091 & \textbf{0.950} & 0.716 \\
\bottomrule
\end{tabular}
\end{table*}

\subsubsection{Concept Localization and Attribution}

The locator comparison examines whether attribution-based word selection provides information beyond random selection or entity detection. For each response, IG attributes the teacher-forced response score $\Phi_i=\sum_{t=1}^{T_i}\log p_\theta(r_{i,t}\mid q_i,r_{i,<t})$, consistent with the definition in Section~\ref{headings}. Prompt embeddings remain fixed, while answer embeddings are interpolated from the padding-token baseline to the original response embeddings. We use 32 FP32 interpolation steps. When a word spans multiple tokens, their attributions are merged into a word-level score. Positive scores are then normalized within each answer and aggregated across questions for each author.

Table~\ref{tab:llm_locator} compares IG with random words, spaCy NER, and model self-redaction. All methods also mask the author name and use the same fine-tuning procedure. Mask ratio is the fraction of answer words that are masked. Gold coverage measures the fraction of gold-answer words covered by the selected words, while precision measures the fraction of selected words that also appear in the gold answers. These are word-selection statistics computed before fine-tuning. Overlap is the fraction of IG-selected words that are also selected by another locator. These statistics are averaged over questions. Cov.~$p_{10}$ denotes the tenth percentile of fact-level coverage across 40 author--relation pairs, with questions for the same pair aggregated before computing the percentile.

All locators satisfy the appearance constraints. IG achieves the highest mean gold coverage on Llama-2 and Qwen2.5-7B and ties with random words on Vicuna. It also achieves the highest lower-tail coverage on Llama-2 and Qwen2.5-7B, with Cov.~$p_{10}=0.179$ and 0.150, respectively. Model self-redaction provides the highest precision, while IG provides broader coverage of gold-answer content. These results motivate IG as the default locator for its coverage of answer content and its direct connection to the model's response likelihood, while NER and model self-redaction provide effective alternatives under the appearance constraints.

\begin{figure*}[tb]
\centering
\subfigure[Llama-2-7B.]{
\label{fig:cov_llama}
\includegraphics[width=0.31\textwidth]{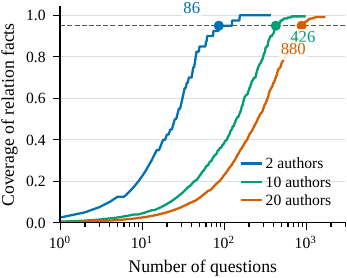}}
\hfill
\subfigure[Vicuna-7B.]{
\label{fig:cov_vicuna}
\includegraphics[width=0.31\textwidth]{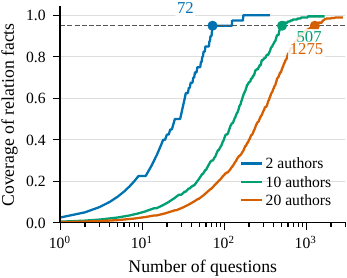}}
\hfill
\subfigure[Qwen2.5-7B.]{
\label{fig:cov_qwen}
\includegraphics[width=0.31\textwidth]{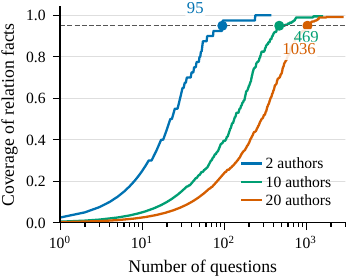}}
\caption{Relation-fact coverage for 2/10/20 forgotten authors. Dashed line: 95\%; markers: smallest counts reaching it. The question-count axis is logarithmic.}
\label{fig:coverage}
\end{figure*}

We also examine the numerical accuracy of IG. With the padding-token baseline, the completeness error is 81.2 at 32 interpolation steps and decreases to 34.2 at 128 steps. A zero-embedding baseline gives a substantially larger error of 360.3 at 32 steps. We use the padding-token baseline with 32 steps as a computationally economical approximation; increasing the number of steps improves numerical completeness.

\begin{table}[tb]
\footnotesize
\setlength{\tabcolsep}{2pt}
\centering
\caption{Answer-target ablation for two forgotten authors. Masked targets replace selected words with \texttt{***}; refusal targets replace the entire answer. Jaccard measures word-set overlap between the constructed training target and the original answer, before fine-tuning. Shading denotes masking; bold marks the best utility and overlap values within each model. $H$ is a count.}
\label{tab:llm_target}
\begin{tabular}{@{}ll*{4}{>{\centering\arraybackslash}p{0.85cm}}*{2}{>{\centering\arraybackslash}p{0.90cm}}@{}}
\toprule
Model & Target & \shortstack{$A_o$\\$\downarrow$} & \shortstack{$A_p$\\$\downarrow$} & \shortstack{$H$\\$\downarrow$} & \shortstack{$R$\\$\uparrow$} & \shortstack{MMLU\\$\uparrow$} & \shortstack{Jaccard\\$\uparrow$} \\
\midrule
\multirow{2}{*}{Llama-2-7B} & \cellcolor{oursblue}Masked & \cellcolor{oursblue}0.000 & \cellcolor{oursblue}0.000 & \cellcolor{oursblue}0.000 & \cellcolor{oursblue}\textbf{0.820} & \cellcolor{oursblue}\textbf{0.455} & \cellcolor{oursblue}\textbf{0.742} \\
 & Refusal & 0.000 & 0.000 & 0.000 & 0.723 & \textbf{0.455} & 0.007 \\
\multirow{2}{*}{Vicuna-7B} & \cellcolor{oursblue}Masked & \cellcolor{oursblue}0.000 & \cellcolor{oursblue}0.000 & \cellcolor{oursblue}0.000 & \cellcolor{oursblue}0.535 & \cellcolor{oursblue}\textbf{0.485} & \cellcolor{oursblue}\textbf{0.764} \\
 & Refusal & 0.000 & 0.000 & 0.000 & \textbf{0.538} & 0.480 & 0.008 \\
\multirow{2}{*}{Qwen2.5-7B} & \cellcolor{oursblue}Masked & \cellcolor{oursblue}0.000 & \cellcolor{oursblue}0.000 & \cellcolor{oursblue}0.000 & \cellcolor{oursblue}\textbf{0.522} & \cellcolor{oursblue}0.630 & \cellcolor{oursblue}\textbf{0.745} \\
 & Refusal & 0.000 & 0.000 & 0.000 & 0.502 & \textbf{0.645} & 0.008 \\
\bottomrule
\end{tabular}\end{table}

\subsubsection{Answer-Target Design}

Both local masking and whole-answer refusal achieve $A_o=A_p=H=0$ in the two-author setting (Table~\ref{tab:llm_target}). Local masking gives higher retained-answer utility $R$ on Llama-2 (0.820 versus 0.723) and Qwen2.5-7B (0.522 versus 0.502), with comparable results on Vicuna (0.535 versus 0.538). Before fine-tuning, masked training targets have a Jaccard word-set overlap of 0.742--0.764 with the original answers, compared with 0.007--0.008 for refusal targets. This measures how much original wording the constructed targets preserve. Overall, local masking meets the same name-appearance constraints while preserving more of the original wording in the training targets and maintaining comparable or higher retained-answer utility.

\subsubsection{Question-Suite Construction and Coverage}

The question suite covers multiple facts and alternative phrasings of each fact. We combine original TOFU questions, official paraphrases, five templates for each of 20 relations, and model-generated paraphrases. Question suites contain up to 2,760 questions, with balanced representation across authors. The relations include birthplace, family, and publications in addition to self-introduction. The model's answer to each candidate question is compared with its answer to the corresponding original question. For candidates whose answers do not explicitly mention the expected author, we require ROUGE-L $\geq0.4$ and content recall $\geq0.5$. Author-name consistency is checked for all candidates to reject conflicting entities. This construction balances paraphrase diversity with relation-level factual coverage.

Coverage is measured over distinct author--relation pairs reached by the first $n$ questions. Reaching 95\% coverage requires 72--95 questions for 2 authors, 426--507 for 10 authors, and 880--1275 for 20 authors (Fig.~\ref{fig:coverage}). The required number grows with the number of author--relation pairs as the forgetting scope increases. These curves provide a direct criterion for selecting a question-suite size that covers the target relations across different forgetting scales.

\subsubsection{Efficiency Analysis}

\begin{table}[htb]
\footnotesize
\setlength{\tabcolsep}{3pt}
\centering
\caption{Runtime breakdown for two-author LLM unlearning (seconds).}
\begin{tabular}{@{}lccc@{}}
\toprule
Stage & Llama-2-7B & Vicuna-7B & Qwen2.5-7B \\
\midrule
Attribution stage & 486 & 257 & 238 \\
Construction stage & $<$1 & $<$1 & $<$1 \\
Unlearning stage & 1062 & 736 & 700 \\
\bottomrule
\end{tabular}
\label{tab:time_llm}
\end{table}

Table~\ref{tab:time_llm} reports the runtime breakdown for two-author unlearning. The attribution stage takes 486 s for Llama-2-7B, 257 s for Vicuna-7B, and 238 s for Qwen2.5-7B. The construction stage requires less than one second for all three models. The unlearning stage takes 1062 s, 736 s, and 700 s, respectively. The results show that runtime is concentrated in attribution and model updating, while poisoned-answer construction adds negligible overhead.

\section{Conclusion}

In this paper, we proposed a concept-guided poisoning unlearning framework for complex data, in which human-interpretable concepts connect the forgetting target to the model representations that produce the target behavior. For image classification, our method uses PCBM-based concept inference and concept localization to identify and modify semantic evidence for a target class, while for LLMs, it uses multi-question analysis and Integrated Gradients to identify sensitive response content and construct locally masked training targets. Experiments on image classification and LLM unlearning tasks show that the proposed method achieves effective unlearning of target information while largely preserving retained model utility across different classes, datasets, models, and forgetting scales. The analyses further show that concept inference provides a more targeted basis for constructing poisoned samples, while training on the modified data weakens the original mapping between target concepts and model outputs. Overall, our work demonstrates that concept-level semantic information can provide a practical interface between unlearning requests and controlled model updates, and offers a unified way to apply poisoning-based unlearning to both image classifiers and LLMs.

\bibliographystyle{IEEEtran}
\bibliography{ref}

\begin{IEEEbiography}
[{\includegraphics[width=1in,height=1.25in,clip,keepaspectratio]{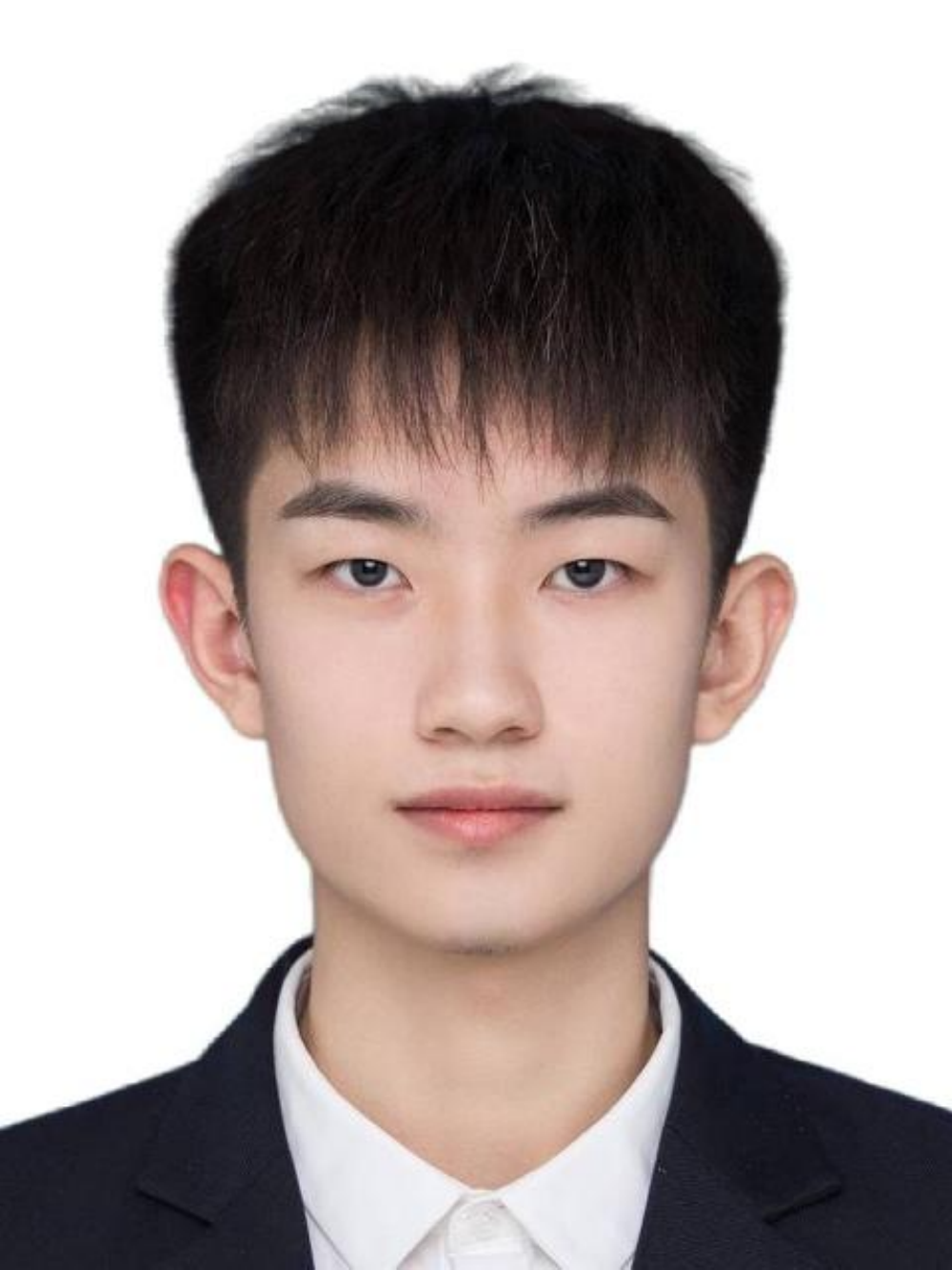}}]{Wenhan Chang} received his B.Eng. degree in 2022 and M.Eng. degree in 2025 from China University of Geosciences, Wuhan, China. He is currently pursuing a Ph.D. degree at the School of Information Engineering, Zhongnan University of Economics and Law, China. His research interests include security and privacy preservation in deep learning.
\end{IEEEbiography}

\begin{IEEEbiography}
[{\includegraphics[width=1in,height=1.25in,clip,keepaspectratio]{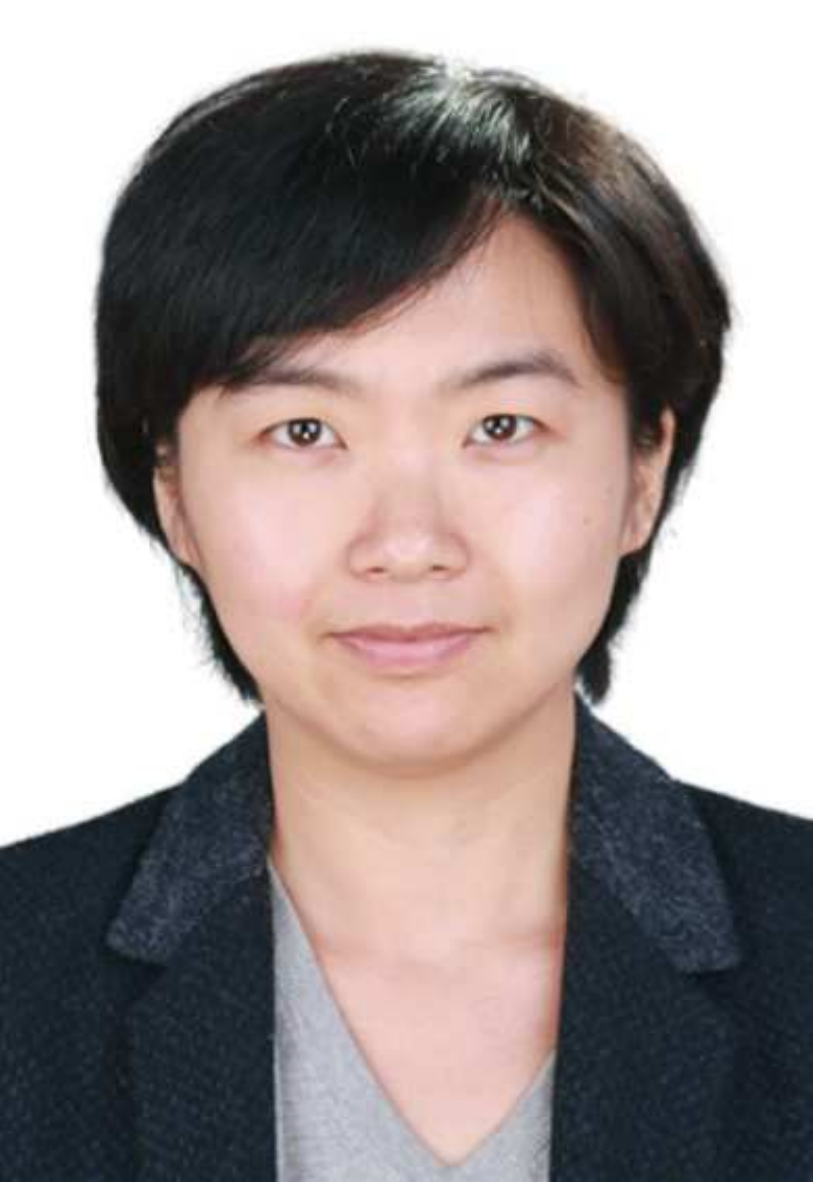}}]
{Tianqing Zhu} received her B.Eng. and M.Eng. degrees from Wuhan University, China, in 2000 and 2004, respectively, and her Ph.D. degree in computer science from Deakin University, Australia, in 2014.
She is currently the Dean of the Faculty of Data Science at the City University of Macau, Macau SAR, China. Prior to that, she was a lecturer with the School of Information Technology at Deakin University and later an associate professor at the University of Technology Sydney. Her research interests include privacy preservation, AI security, and network security.
\end{IEEEbiography}

\begin{IEEEbiography}[{\includegraphics[width=1in,height=1.3in,clip,keepaspectratio]{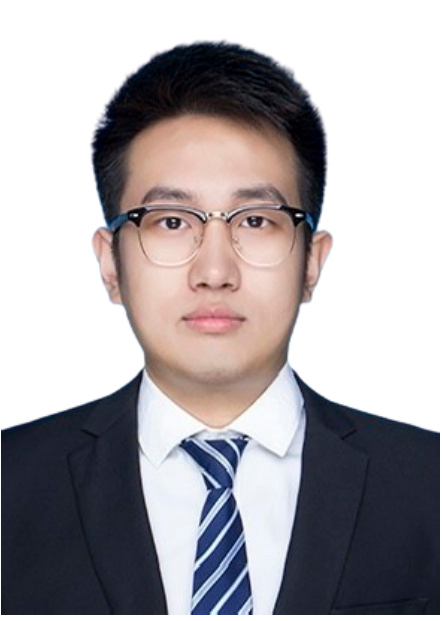}}]{Heng Xu} received the B.Eng. and M.Eng. degrees from China University of Geosciences, Wuhan, China, in 2018 and 2021, respectively, and the Ph.D. degree in computer science from the University of Technology Sydney, Australia, in 2025. He is currently an Assistant Professor with the Faculty of Data Science at the City University of Macau. His research interests are machine unlearning and AI security.
\end{IEEEbiography}

\begin{IEEEbiography}
[{\includegraphics[width=1in,height=1.25in,clip,keepaspectratio]{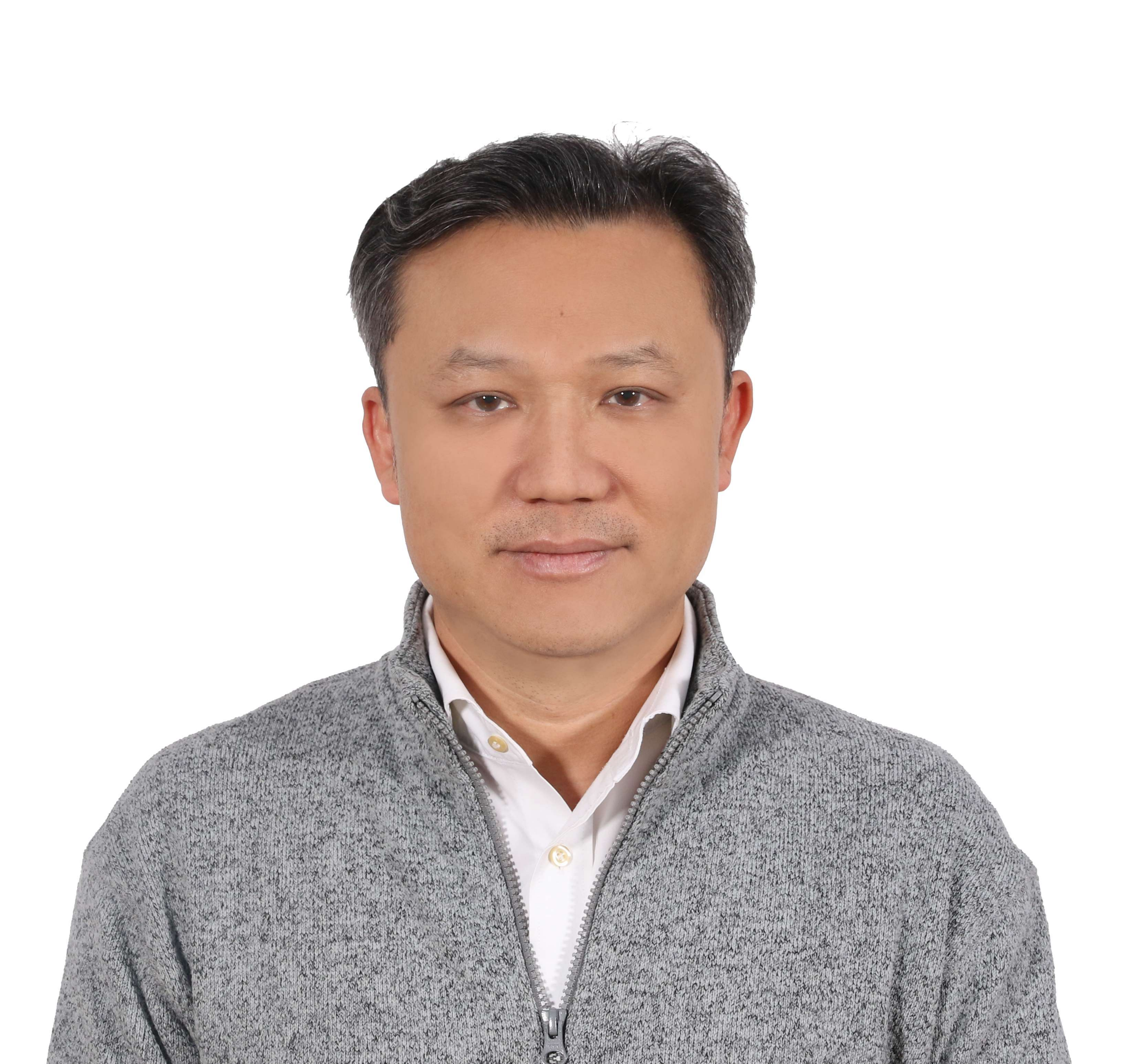}}]
{Wenjian Liu} received the B.S., M.S., and Ph.D. degrees from the School of Electrical and Information Engineering, South China University of Technology, Guangzhou, China. He is currently an Associate Professor of the Faculty of Data Science, City University of Macau. His research interests focus on the application of AI across interdisciplinary fields such as intelligent healthcare, generative artificial intelligence, AI security, and smart cities.
\end{IEEEbiography}

\begin{IEEEbiography}
[{\includegraphics[width=1.1in,height=1.25in,clip,keepaspectratio]{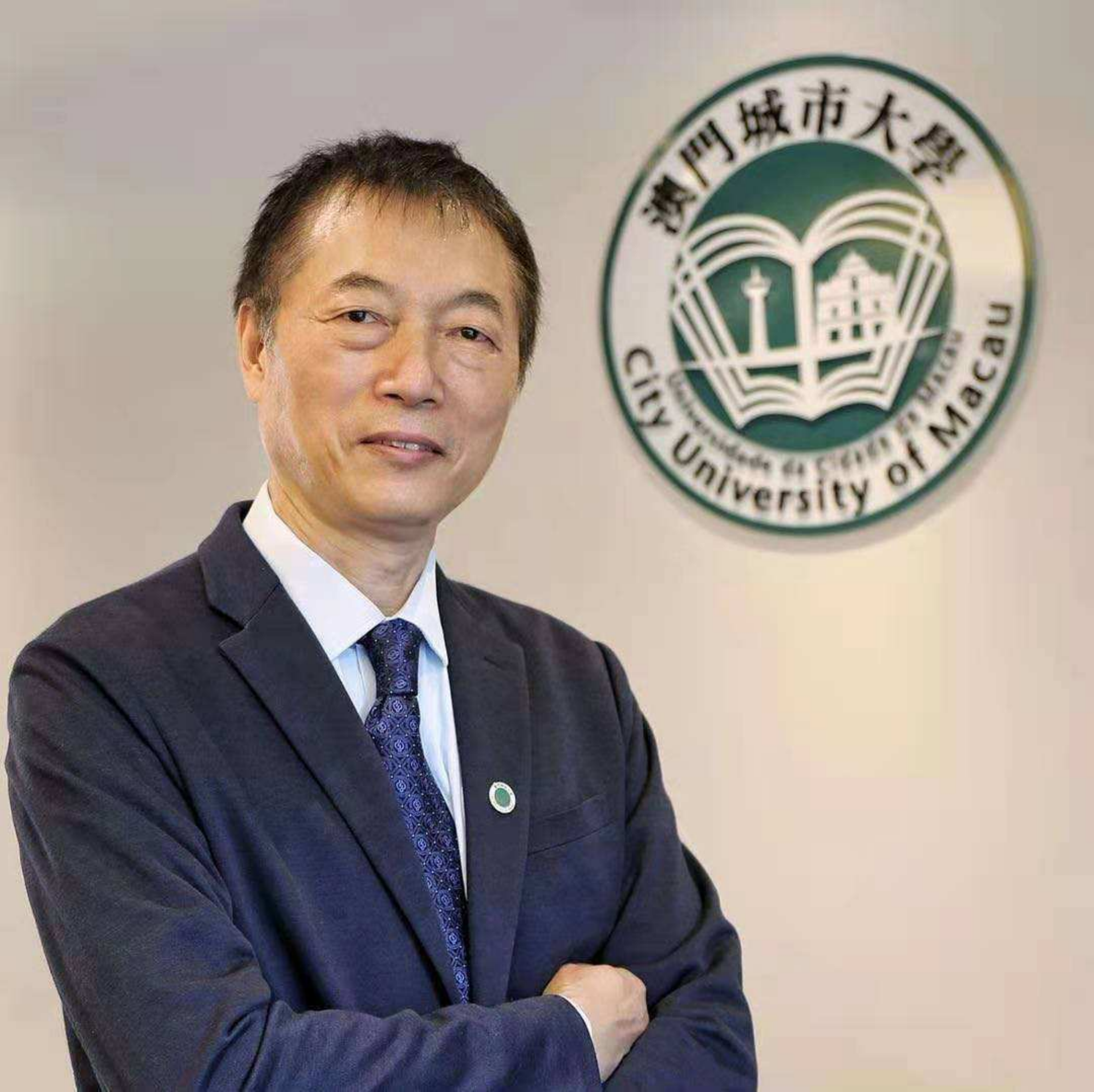}}]
{Wanlei Zhou} received the B.Eng. and M.Eng. degrees from Harbin Institute of Technology, Harbin, China, in 1982 and 1984, respectively, and the Ph.D. degree from The Australian National University, Canberra, Australia, in 1991, all in computer science and engineering. He also received a D.Sc. degree from Deakin University in 2002. He is currently the Vice Rector and Dean of the Institute of Data Science at the City University of Macau, Macau SAR, China. He has authored or coauthored more than 400 papers in refereed international journals and conference proceedings, including many articles in IEEE Transactions and journals. His research interests include security and privacy, parallel and distributed systems, and e-learning.
\end{IEEEbiography}

\vfill

\end{document}